\pdfoutput=1
\documentclass[twoside]{article}

\usepackage[accepted]{aistats2024}
\usepackage[round]{natbib}

\usepackage{ISG_style_reduced}

\usepackage[utf8]{inputenc} 
\usepackage[T1]{fontenc}    
\usepackage[hidelinks]{hyperref}       
\usepackage{url}            
\usepackage{booktabs}       
\usepackage{amsfonts,amssymb,amsmath}       
\usepackage{nicefrac}       
\usepackage{microtype}      
\usepackage{amssymb}
\usepackage{physics}        
\usepackage{graphicx}
\usepackage{subcaption}
\usepackage{bbm}
\usepackage{dsfont}
\usepackage{cite}
\usepackage{flushend}
\usepackage{float}
\usepackage[dvipsnames]{xcolor}
\usepackage{algorithm}
\usepackage{algpseudocode}
\usepackage{mathtools}
\usepackage{aligned-overset}
\usepackage{comment}
\usepackage{tablefootnote}
\usepackage{enumitem}
\usepackage{thm-restate}
\usepackage{soul}

\DeclareMathOperator{\image}{im}

\DeclareMathOperator{\sgn}{sgn}

\DeclareMathOperator{\arctanh}{arctanh}

\usepackage{centernot}
\def\ci{\perp\!\!\!\perp}
\def\notci{\centernot\perp\!\!\!\perp}

\makeatletter
\let\assumption\@undefined
\let\endassumption\@undefined
\newtheorem{assumption}{Assumption}
\let\example\@undefined
\let\endexample\@undefined
\newtheorem{example}{Example} 
\let\theorem\@undefined
\let\endtheorem\@undefined
\newtheorem{theorem}{Theorem}
\let\lemma\@undefined
\let\endlemma\@undefined
\newtheorem{lemma}{Lemma} 
\let\proposition\@undefined
\let\endproposition\@undefined
\newtheorem{proposition}{Proposition} 
\let\remark\@undefined
\let\endremark\@undefined
\newtheorem{remark}{Remark}
\let\corollary\@undefined
\let\endcorollary\@undefined
\newtheorem{corollary}{Corollary}
\let\definition\@undefined
\let\enddefinition\@undefined
\newtheorem{definition}{Definition}
\let\conjecture\@undefined
\let\endconjecture\@undefined
\newtheorem{conjecture}{Conjecture}
\let\axiom\@undefined
\let\endaxiom\@undefined
\newtheorem{axiom}{Axiom}
\makeatother

\begin{document}

\runningtitle{General Identifiability and Achievability for Causal Representation Learning}

\runningauthor{Var{\i}c{\i}, Acart{\"u}rk, Shanmugam, Tajer}

\twocolumn[
\aistatstitle{General Identifiability and Achievability for \\ Causal Representation Learning}
\aistatsauthor{Burak Var\i c\i$^1$ \And Emre Acart\"urk$^1$  \And Karthikeyan Shanmugam$^2$ \And Ali Tajer$^1$ }\vspace{0.1in}
\aistatsaddress{$^1$Rensselaer Polytechnic Institute \And $^2$Google Research India}
]

\begin{abstract}
  This paper focuses on causal representation learning (CRL) under a general nonparametric latent causal model and a general transformation model that maps the latent data to the observational data. It establishes \textbf{identifiability} and \textbf{achievability} results using two hard \textbf{uncoupled} interventions per node in the latent causal graph. Notably, one does not know which pair of intervention environments have the same node intervened (hence, uncoupled). For identifiability, the paper establishes that perfect recovery of the latent causal model and variables is guaranteed under uncoupled interventions. For achievability, an algorithm is designed that uses observational and interventional data and recovers the latent causal model and variables with provable guarantees. This algorithm leverages score variations across different environments to estimate the inverse of the transformer and, subsequently, the latent variables. The analysis, additionally, recovers the identifiability result for two hard \textbf{coupled} interventions, that is when metadata about the pair of environments that have the same node intervened is known. This paper also shows that when observational data is available, additional faithfulness assumptions that are adopted by the existing literature are unnecessary.
\end{abstract}

\section{INTRODUCTION}\label{sec:introduction}
Consider a causal graph $\mcG$ with $n$ nodes generating \emph{causal} random variables $Z \triangleq  [Z_1,\dots,Z_n]^{\top}$. These random variables are transformed by a function $g: \R^n \to \R^d$ to generate the $d$-dimensional \emph{observed} random variables $X \triangleq [X_1,\dots,X_d]^{\top}$ according to:
\begin{equation}
    X = g(Z) \  , \label{eq:data-generation-process-intro}
\end{equation}
where $\image(g)\triangleq \mcX \subseteq \R^d$. Causal representation learning (CRL) is the process of using the data $X$ and recovering (i)~\textbf{the causal structure} $\mcG$ and (ii)~\textbf{the latent causal variables} $Z$. When interventions are viable, the process is referred to as CRL from interventions. Achieving these implicitly involves another objective of recovering the unknown transformation $g$ as well. Addressing CRL consists of two central questions:
\begin{itemize}[leftmargin=1em]
    \item {\bf Identifiability,} which refers to determining the necessary and sufficient conditions under which $\mcG$ and $Z$ can be recovered. Literature on CRL from interventions commonly assumes that interventional and observational distributions are sufficiently different and inherit faithfulness assumption from causal discovery literature. Note that identifiability can be non-constructive without specifying how to recover $\mcG$ and $Z$. 
    \item {\bf Achievability}, which complements identifiability and pertains to designing algorithms that can recover $\mcG$ and $Z$ while maintaining identifiability guarantees. Achievability hinges on forming reliable estimates for the transformation $g$.
\end{itemize}
This paper provides both identifiability and achievability results for CRL under stochastic \emph{hard} interventions when (i) the transformation $g$ can be any function (linear or non-linear) that is a diffeomorphism (i.e., bijective such that both $g$ and $g^{-1}$ are continuously differentiable) onto its image, and (ii) the causal relationships among elements of $Z$ take any arbitrary form (linear or non-linear). Specifically, our main contributions are:

\begin{table*}[ht]
  \centering
  \caption{\small Comparison of the results to prior studies in different settings. Only the main results from the papers that aim \emph{both} DAG and latent recovery are listed. See Section~\ref{sec:problem} for exact definitions of perfect DAG and latent recovery. Additional assumptions ($^*$:interventional discrepancy and $^{**}$: faithfulness) are discussed in Section~\ref{sec:identifiability}.}
  \scalebox{0.8}{
  \begin{tabular}{cccccccc}
      \toprule
      Work & Transform & Latent Model & Obs. Data & Interv. Env. (per node)  & DAG recovery & Latent recovery \\
      \hline 
      \citep{squires2023linear} & Linear & Lin. Gaussian & Yes & 1 Soft & impossibility & impossibility  \\
       & Linear & Lin. Gaussian & Yes & 1 Hard & Yes & Yes \\
          \midrule 
      \citep{ahuja2023interventional} & Polynomial  & General & Yes & 1 \emph{do} & Yes & Yes \\
      & Polynomial & Bounded RV & Yes & 1 Soft & Yes & Yes \\
      \hline
      \citep{varici2023score} & Linear & Non-linear
          & Yes & 1 Soft & Yes & Mixing \\
      & Linear & Non-linear
          & Yes & 1 Hard & Yes & Yes \\        \bottomrule
      \citep{buchholz2023learning} & General & Lin. Gaussian & Yes & 1 Hard & Yes & Yes \\ \bottomrule
      \citep{zhang2023identifiability} & Polynomial & Non-linear & Yes & 1 Soft & Yes & Yes  \\ \bottomrule
      \citep{vonkugelgen2023twohard} & General & General &  No & 2 coupled Hard$^*$ & Yes$^{**}$ & Yes \\ \bottomrule
    \textbf{Theorem~\ref{th:two-hard-uncoupled}} & General & General & Yes & 2 uncoupled Hard$^*$  & Yes & Yes \\ 
      \textbf{Theorem~\ref{th:two-hard-coupled}} & General & General & Yes & 2 coupled Hard$^*$& Yes & Yes \\ 
      \textbf{Theorem~\ref{th:faithfulness}} & General & General & No & 2 coupled  Hard$^*$ & Yes$^{**}$ & Yes \\
      \bottomrule
  \end{tabular}}
  \label{tab:related-comp}
\end{table*} 

\begin{itemize}[leftmargin=1em]
    \item On identifiability, we show that two \emph{uncoupled} hard interventions per node suffice to guarantee perfect nonparametric identifiability (up to permutation and element-wise transforms). Specifically, we assume the learner does not know which pair of environments intervene on the same node, hence, uncoupled. 

     \item On achievability, we design the first provably correct algorithm that recovers $\mcG$ and $Z$ under a general transformation and a causal model. This algorithm leverages the variations of the score functions under interventions, and is referred to as {\bf G}eneralized {\bf S}core-based {\bf Ca}usal {\bf L}atent {\bf E}stimation via {\bf I}nterventions (GSCALE-I).

    \item While establishing identifiability results, we show that faithfulness assumptions are not required when observational data is available in contrast to recent results in the literature that require faithfulness assumptions.

\end{itemize}

\subsection{Background}
Learning latent representations from high-dimensional observational data is an important task in machine learning which can significantly improve the generalization and robustness of the models \citep{bengio2013representation}. To that end, disentangled representation learning aims to infer a latent representation such that each latent variable corresponds to a meaningful component and the latent variables are statistically independent, similar to independent component analysis (ICA). In a wide range of domains, however, the latent variables are not independent and instead are related causally.

Causal representation learning aims to build realistic models with a causal understanding of the world \citep{scholkopf2021toward}. In CRL, the generation of each latent variable is governed by a causal mechanism, i.e., a conditional probability kernel. \emph{Identifiability} is known to be impossible without additional supervision or sufficient statistical diversity among the samples of the observed data $X$. As shown in~\citep{hyvarinen1999nonlinear,locatello2019challenging}, this is the case even for the simpler settings in which the latent variables are statistically independent (i.e., graph $\mcG$ has no edges). Hence, identifiability guarantees critically hinge on the additional information on the data generation process and the observed data. A central problem in the literature is investigating the identifiability of CRL when performed in conjunction with \emph{interventions}~\citep{scholkopf2021toward}. Specifically, leveraging the modularity property of causal models \citep{pearl2009causality}, upon an intervention, \emph{only} the manipulated causal mechanisms related to intervention targets change. This creates proper variations in the observed data, which facilitates identifiability. 

\subsection{Related Work} 
In this paper, we address identifiability and achievability in CRL given different interventional environments in which the interventions act on the latent space. The recent studies have been almost entirely focused on the identifiability results. We provide an overview of these existing results and discuss the closely related literature that investigates CRL from interventional data, with the main results summarized in Table~\ref{tab:related-comp}. We defer the discussion on other approaches to CRL with stronger supervision signals, e.g., counterfactual image pairs and temporal data, to Appendix~\ref{appendix:related}.

The recent studies most closely related to the scope of this paper are~\citep{varici2023score}~and \citep{vonkugelgen2023twohard}. \citet{varici2023score} establish an inherent connection between score functions and CRL and leverages that connection to design a score-based CRL framework. Specifically, using \emph{one intervention per node} under a non-linear causal model and a linear transformation,~\citep{varici2023score} provides both identifiability and achievability results. We have three major distinctions from~\citep{varici2023score} in settings by assuming nonparametric choices of transformations $g$, a general latent causal model, and using two hard interventions. 

\citet{vonkugelgen2023twohard} consider nonparametric models for $g$ and the causal relationships and shows that two \emph{coupled} hard interventions per node suffice for identifiability. We have two major differences. First, we assume uncoupled interventional environments, whereas~\citep{vonkugelgen2023twohard} focuses on coupled environments. Secondly, the approach in~\citep{vonkugelgen2023twohard} focuses mainly on identifiability (e.g., no algorithm for recovery of the latent variables), whereas we address both identifiability and achievability. 

Other related studies that focus on the \emph{parametric} settings include~\citep{squires2023linear,ahuja2023interventional,zhang2023identifiability,buchholz2023learning}. Specifically, \citet{squires2023linear} consider linear causal models and proves identifiability under hard interventions and the impossibility of identifiability under soft interventions. \citet{ahuja2023interventional} consider a polynomial transform and shows that it can be reduced to an affine transform by an autoencoding process and proves identifiability under \emph{do} interventions or soft interventions on bounded latent variables. \citet{zhang2023identifiability} consider a polynomial transform under non-linear causal models and prove identifiability under soft interventions. Finally, \citet{buchholz2023learning} focus on linear Gaussian causal models and extend the results in \citep{squires2023linear} to prove identifiability for general transforms. For \emph{nonparametric} settings, \citet{jiang2023learning} focus on identifying the latent DAG without recovering latent variables, and show that a restricted class of DAGs can be recovered. \citet{liang2023causal} assume that the latent DAG is already known and recover the latent variables under hard interventions.

\section{PROBLEM SETTING}\label{sec:problem}

\noindent\textbf{Notations.} For a vector $a\in\R^{m}$, the $i$-the entry is denoted by $a_i$. Matrices are denoted by bold upper-case letters, e.g., $\bA$, where $\bA_i$ denotes the $i$-th row of $\bA$ and $\bA_{i,j}$ denotes the entry at row $i$ and column $j$. We denote the indicator function by $\mathds{1}$, and for a matrix $\bA\in\R^{m\times n}$, we use the convention that $\mathds{1}\{\bA\}\in\{0,1\}^{m\times n}$, where the entries are specified by $[\mathds{1}\{\bA\}]_{i,j}=\mathds{1}\{\bA_{i,j}\neq 0\}$. For a positive integer $n\in\N$, we define $[n] \triangleq \{1,\dots,n\}$.  The permutation matrix associated with any permutation $\pi$ of $[n]$ is denoted by $\bP_{\pi}$, i.e., $[\pi_1 \;\;  \pi_2 \; \dots   \; \pi_n]^\top=\bP_{\pi}\cdot[1 \;\; 2 \; \dots \;  n]^\top$. The $n$-dimensional identity matrix is denoted by $\bI_{n \times n}$, and the Hadamard product is denoted by $\odot$. Given a function $f: \R^r \to \R^s$ that has first-order partial derivatives on $\R^r$, we denote the Jacobian of $f$ at $z\in\R^s$ by $J_{f}(z)$. We use $\image(f)$ to denote the image of $f$.

\noindent\textbf{Latent causal structure.} Consider latent causal random variables $Z \triangleq  [Z_1,\dots,Z_n]^{\top}$. An \emph{unknown} transformation $g: \R^n \to \R^d$ generates the observable random variables $X \triangleq [X_1,\dots,X_d]^{\top}$ from the latent variables $Z$ according to:
\begin{equation}
    X = g(Z) \  . \label{eq:data-generation-process}
\end{equation}
We assume that $d \geq n$, and transformation $g$ is continuously differentiable and a diffeomorphism onto its image (otherwise, identifiability is ill-posed). We denote the image of $g$ by $\mcX \triangleq  \image(g)\subseteq \R^d$. The probability density functions (pdfs) of $Z$ and $X$ are denoted by $p$ and $p_X$, respectively. We assume that $p$ is absolutely continuous with respect to the $n$-dimensional Lebesgue measure. Subsequently, $p_X$, which is defined on the image manifold $\image(g)$, is absolutely continuous with respect to the $n$-dimensional Hausdorff measure rather than $d$-dimensional Lebesgue measure. The distribution of latent variables $Z$ factorizes with respect to a DAG that consists of $n$ nodes and is denoted by $\mcG$. Node $i\in[n]$ of $\mcG$ represents $Z_i$ and $p$ factorizes according to:
\begin{equation}
    p(z) = \prod_{i=1}^{n} p_i(z_i \med z_{\Pa(i)}) \ , \label{eq:pz_factorized}
\end{equation}
where $\Pa(i)$ denotes the set of parents of node $i$ and $p_i(z_i \med z_{\Pa(i)})$ is the conditional pdf of $z_i$ given the variables of its parents. We also define $\overline{\Pa}(i)\triangleq \Pa(i) \cup \{i\}$, and use $\Ch(i)$ to denote the children of node $i$. Based on the modularity property, a change in the causal mechanism of node~$i$ does not affect those of the other nodes. We also assume that all conditional pdfs $\{p_i(z_i \mid z_{\Pa(i)}) : i \in[n]\}$ are continuously differentiable with respect to all $z$ variables and $p(z) \neq 0$ for all $z \in \R^n$. \looseness=-1

\noindent\textbf{Score functions.} The \emph{score function} associated with a pdf is defined as the gradient of its logarithm. The score function associated with $p$ is denoted by 
\begin{equation}
    s(z) \triangleq \nabla_z \log p(z) \ . \label{eq:sz-def}
\end{equation}
Noting the connection $X=g(Z)$, the density of $X$ under $\mcE^0$, denoted by $p_X$, is supported on an $n$-dimensional manifold $\mcX$ \emph{embedded} in $\R^{d}$. Hence, specifying the score function of $X$ requires notions from differential geometry. For this purpose, we denote the tangent space of manifold $\mcX$ at point $x \in \mcX$ by $T_{x} \mcX$. Furthermore, given a function $f \colon \mcX \to \R$, denote its directional derivative at point $x \in \mcX$ along a tangent vector $v \in T_{x} \mcX$ by $D_{v} f(x)$. The differential of $f$ at point $x \in \mcX$, denoted by $\dd f_x$, is the linear operator mapping tangent vector $v \in T_{x} \mcX$ to $D_{v} f(x)$ \citep[p.~57]{simon2014introduction}, i.e.,
\begin{equation}
    \dd f_{x} \colon T_{x} \mcX \ni v \mapsto D_{v} f(x) \in \R \ . \label{eq:differential-def}
\end{equation}
Let $\bB \in \R^{d \times n}$ be a matrix for which the columns of $\bB$ form an orthonormal basis for $T_{x} \mcX$. Denote the directional derivative of $f$ along the $i$-th column of $\bB$ by $D_i f$ for all $i \in [n]$. Then, the differential operator can be expressed by the vector
\begin{equation}
    D f(x) \triangleq \bB \cdot \big[ D_1 f(x) \dots D_n f(x) \big]^{\top} \in \R^{d} \ , \label{eq:Df-defn}
\end{equation}
such that $\dd f_x(v) = v^{\top} \cdot D f(x)$ for all $x \in \mcX$ and $v \in T_{x} \mcX$.
Note that the differential operator $\dd f$ is a generalization of the gradient. Hence, we can generalize the definition of the score function using the differential operator by setting $f$ to the logarithm of pdf. Therefore, the score function of $X$ under $\mcE^0$ is specified as follows:
\begin{equation}
    s_{X}(x) \triangleq D \log p_{X}(x) \ , \qquad \forall x \in \mcX \ . \label{eq:sx-def}
\end{equation}
\noindent\textbf{Intervention models.}
For each node $i \in [n]$, besides the observational mechanism specified by $p_i(z_i \med z_{\Pa(i)})$, we assume that there exist two hard interventional mechanisms specified by $q_i(z_i)$ and $\tilde q_i(z_i)$. 
We assume \emph{interventional discrepancy}~\citep{liang2023causal} among the distributions.
\begin{definition}[Interventional discrepancy]
Two mechanisms with pdfs $p,q:\R \to \R$ satisfy \emph{interventional discrepancy} if 
\begin{equation}\label{eq:int-discrepancy}
    \frac{\partial}{\partial u} \frac{p(u)}{q(u)} \neq 0 \ , \quad {\forall u\in \R \setminus \mcT} \ , 
\end{equation}
where $\mcT$ is a null set (i.e., has Lebesgue measure zero). 
\end{definition}
We note that~\citep{liang2023causal} shows that for identifiability in the single atomic hard intervention per node setting, even when the latent graph $\mcG$ is known, it is necessary to have an interventional discrepancy between observational distribution $p_i$ and interventional distribution $q_i$, for all $z_{\Pa(i)}\in\R^{|\Pa(i)|}$.

\noindent\textbf{Interventional environments.}
We consider two sets of atomic interventional environments denoted by $\mcE \triangleq \{\mcE^{m} : m \in [n]\}$ and $\tilde \mcE \triangleq \{\tilde \mcE^{m} : m \in [n]\}$, for which a single node is intervened in each environment. We denote the node intervened in environment $\mcE^m$ by $I^m \in [n]$, and similarly denote the intervened node in $\tilde \mcE^m$ by $\tilde I^m$. We assume that node-environment pairs are \emph{unspecified}, i.e.,  the ordered intervention sets  $\mcI\triangleq (I^1,\dots,I^n)$ and $\tilde \mcI \triangleq (\tilde I^1,\dots,\tilde I^n)$ are two \emph{unknown} permutations of~$[n]$. We also adopt the convention that $I^{0} = \emptyset$, and define $\mcE^{0}$ as the \emph{observational} environment.  Next, we define the notion of coupling between the environment sets $\mcE$ and $\tilde \mcE$.
\begin{definition}[Coupled/Uncoupled environments]
The two environment sets $\mcE$ and $\tilde\mcE$ are said to be \emph{coupled} if for the unknown permutations $\mcI$ and $\tilde \mcI$ we know that $\mcI=\tilde\mcI$, i.e., the same node is intervened in environments $\mcE^i$ and $\tilde \mcE^i$. The two environment sets are said to be \emph{uncoupled} if $\tilde\mcI$ is an unknown permutation of $\mcI$. 
\end{definition}
We denote the pdfs of $Z$ under the hard interventions in environments $\mcE^m$ and $\tilde\mcE^m$, by $p^m$ and $\tilde p^m$, respectively, which can be factorized as follows for all $m\in[n]$.
\begin{align}
    \mcE^m:  & \;\; p^m(z) = q_\ell(z_{\ell}) \prod_{i \neq \ell} p_i(z_i \med z_{\Pa(i)}) \ , \; \; \ell=I^m \ ,\label{eq:pz_m_factorized_hard} \\
    \tilde \mcE^m: & \;\; \tilde p^m(z) = \tilde q_{\ell}(z_{\ell}) \prod_{i \neq \ell} p_i(z_i \med z_{\Pa(i)})\ , \;   \;  \ell=I^m \ . \label{eq:pz_m_factorized_hard_tilde} 
\end{align}
Hence, the score functions associated with $p^m$ and $\tilde p^m$ are specified as follows.
\begin{equation}
    s^{m}(z) \triangleq \nabla_z \log p^{m}(z) \ , \; \mbox{and} \;\; \tilde s^{m}(z) \triangleq \nabla_z \log \tilde p^{m}(z) \ . \label{eq:sz-defs}
\end{equation}

\noindent\textbf{Statement of the objective.}
The objective of CRL is to use observations $X$ and estimate the true latent variables $Z$ and causal relations among them captured by $\mcG$. We denote a generic estimator of $Z$ given $X$ by $\hat Z(X):\R^d\to\R^n$. We also consider a generic estimate of $\mcG$ denoted by $\hat\mcG$. In order to assess the fidelity of the estimates $\hat Z(X)$ and $\hat\mcG$ with respect to the ground truth $Z$ and $\mcG$, we provide the following identifiability measures. The result in~\citep[Proposition 3.8]{vonkugelgen2023twohard} shows that these identifiability measures are the best one can ensure based on interventional data without more direct forms of supervision, e.g., counterfactual data. 

\begin{definition}[Perfect Identifiability]\label{def:identifiability}
    To formalize perfect identifiability in CRL we define:
        \begin{enumerate}[leftmargin=2em]
           \item \textbf{\em Perfect DAG recovery:} DAG recovery is said to be perfect if $\hat \mcG$ is isomorphic to $\mcG$.
        
            \item \textbf{\em Perfect latent recovery:} Given the estimator $\hat Z(X)$, latent recovery is said to be perfect if $\hat Z(X)$ is an element-wise diffeomorphism of a permutation of $Z$, i.e., there exists a permutation~$\pi$ of~$[n]$ and a set of functions $\{\phi_i: i\in[n]\}$ such that $\phi_i:\R \to \R$ and we have
            \begin{equation}
                \hat Z(X) = \bP_{\pi} \cdot \phi(Z) \ , \quad \forall Z\in\R^n\ ,
            \end{equation}
            where $\phi(Z)\triangleq (\phi_1(Z_1),\dots,\phi_n(Z_n)$.
        \end{enumerate} 
\end{definition}
Recovering the latent causal variables hinges on finding the inverse of $g$ based on the observed data $X$, which in turn facilitates recovering $Z$ via $Z = g^{-1}(X)$, where $g^{-1}$ denotes the inverse of $g$. Throughout the rest of this paper, we refer to $g^{-1}$ as the \emph{true} encoder. To formalize the procedures of estimating $g^{-1}$, we define $\mcH$ as the set of all possible valid encoders, i.e., candidates for $g^{-1}$. A function $h$ can be such a candidate if it is invertible; that is, there exists an associated decoder $h^{-1}$ such that $(h^{-1} \circ h)(X) = X$. Hence,
\begin{align}\label{eq:mcH}
    \mcH \triangleq \{ h: \; &\mcX \to \R^n \;\colon\; \exists h^{-1}: \R^n \to \R^d \;\;  \\ &\mbox{such that }\; \forall X \in \mcX: \; (h^{-1} \circ h)(X) = X \} \ . \notag
\end{align}
\looseness=-1
Next, corresponding to any pair of observation $X$ and valid encoder $h\in\mcH$, we define $\hat Z(X;h)$ as an \emph{auxiliary} estimate of $Z$ generated by applying the valid encoder $h$ on $X$, i.e., for all $h\in\mcH$ and $X\in\mcX$
\begin{equation}\label{eq:encoder}
    \hat Z(X;h) \triangleq h(X) = (h \circ g) (Z) \ . 
\end{equation}
The estimate $\hat Z(X;h)$ inherits its randomness from $X$, and its statistical model is governed by that of~$X$ and the choice of $h$. To emphasize the dependence on $h$, we denote the score functions associated with the pdfs of $\hat Z(X;h)$ under environments $\mcE^{0}$, $\mcE^m$, and $\tilde\mcE^m$, respectively, by $s_{\hat Z}(\cdot ;h)$, $s_{\hat Z}^{m}(\cdot ;h)$, and $\tilde s_{\hat Z}^{m}(\cdot;h)$.

\section{IDENTIFIABILITY AND ACHIEVABILITY RESULTS}\label{sec:identifiability}

In this section, we provide the identifiability and achievability results under different sets of assumptions and interpret them vis-\'a-vis the recent results in the literature. We provide constructive proof for the results by designing CRL algorithms. The details of the CRL algorithm are summarized in Algorithm~\ref{alg:main}, which is presented in Section~\ref{sec:algorithm}. Our main result is the following theorem, which establishes perfect identifiability is possible even when the environments corresponding to the same node are not specified in pairs. That is, not only is it unknown what node is intervened in an environment, additionally the learner also does not know which two environments intervene on the same node. 
\setcounter{theorem}{0}
\begin{theorem}[Uncoupled Environments]\label{th:two-hard-uncoupled}
    Using observational data and interventional data from two \emph{uncoupled} environments for which each pair in $\{p_i, q_i, \tilde q_i\}$ satisfies interventional discrepancy, suffices to \looseness = -1 \vspace{-.05 in}
    \begin{enumerate}[label=(\roman*), leftmargin=1.5em, itemsep=-.05 in]
        \item Identifiability: perfectly recover the latent DAG;
        \item Identifiability: perfectly recover the latent variables; 
        \item Achievability: achieve the above two guarantees (via Algorithm~\ref{alg:main}).
    \end{enumerate}
\end{theorem}
Theorem~\ref{th:two-hard-uncoupled} shows that using observational data enables us to resolve any mismatch between the uncoupled environment sets and shows identifiability in the setting of uncoupled environments. This generalizes the identifiability result of \citep{vonkugelgen2023twohard}, which requires coupled environments. Importantly, Theorem~\ref{th:two-hard-uncoupled} does not require faithfulness whereas \citep{vonkugelgen2023twohard} requires that the estimated latent distribution is faithful to the associated candidate graph for all $h \in \mcH$. Even though a faithfulness assumption does not compromise the identifiability result, it is a strong requirement to verify and poses challenges to devising recovery algorithms. In contrast, we only require observational data, which is generally accessible in practice. Based on this, we design a score-based algorithm, which is presented and discussed in Section~\ref{sec:algorithm}. Next, if the environments are coupled, we prove identifiability under weaker assumptions on the interventional discrepancy.
\begin{theorem}[Coupled Environments]\label{th:two-hard-coupled}
    Using observational data and interventional data from two \emph{uncoupled} environments for which the pair $(p_i, q_i)$ satisfies interventional discrepancy for all $i \in [n]$, suffices to\vspace{-.05 in}
    \begin{enumerate}[label=(\roman*), leftmargin=1.5em, itemsep=-.05 in]
        \item Identifiability: perfectly recover the latent DAG;
        \item Identifiability: perfectly recover the latent variables; 
        \item Achievability: achieve the above two guarantees (via Algorithm~\ref{alg:main}).
    \end{enumerate}
\end{theorem}

In the proof of Theorem~\ref{th:two-hard-coupled}, we show that the advantage of environment coupling is that it renders interventional data sufficient for perfect latent recovery, and the observational data is only used for recovering the graph. We further tighten this result by showing that for DAG recovery, the observational data becomes unnecessary when we have additive noise models and a weak faithfulness condition holds.
\begin{theorem}[No Observational Data]\label{th:faithfulness}
    Using interventional data from two \emph{coupled} environments for which the pair $(q_i,\tilde q_i)$ satisfies interventional discrepancy for all $i \in [n]$, suffices to \vspace{-.05 in}
    \begin{enumerate}[label=(\roman*), leftmargin=2em, itemsep=-.05 in]
        \item Identifiability: perfectly recover the latent DAG if the latent causal model has additive noise, $p(Z)$ is twice differentiable, and it satisfies the adjacency-faithfulness~\footnote{Adjacency-faithfulness is a weaker version of the faithfulness assumption~\citep{ramsey2012adjacency}. It requires that if nodes $i$ and $j$ are adjacent in $\mcG$, then $Z_i$ and $Z_j$ are dependent conditional on any subset of $\big\{Z_\ell : \ell \notin \{i,j\}\big\}$.};
        \item Identifiability: perfectly recover the latent variables  
        \item Achievability: achieve the above two guarantees.
    \end{enumerate}
\end{theorem}

Finally, we remark that the main results in this subsection (i.e., Theorem~\ref{th:two-hard-uncoupled} and \ref{th:two-hard-coupled}) hold for \emph{any} latent causal model. The additive noise model assumption is required only for the DAG recovery part of Theorem~\ref{th:faithfulness}.  

\section{GSCALE-I ALGORITHM}\label{sec:algorithm}

This section serves a two-fold purpose. First, it provides the constructive proof steps for the identifiability results specified in Theorems~\ref{th:two-hard-uncoupled}--\ref{th:two-hard-coupled}. Secondly, it provides achievability via designing an algorithm that has provable guarantees for perfect recovery of the latent variables and latent DAG for any {\em general} class of functions (linear and non-linear). We refer to this algorithm as the {\bf G}eneralized {\bf S}core-based {\bf Ca}usal {\bf L}atent {\bf E}stimation via {\bf I}nterventions (GSCALE-I) algorithm. Analysis of the steps involved are provided in Section~\ref{sec:analysis}.

A key idea of this score-based algorithm is that the changes in the score functions of the latent variables enable us to find reliable estimates for the inverse of transformation $g$, which in turn facilitates estimating $Z$. On the other hand, we do not have access to the latent variables and can compute only the scores of the observed variables $X$. To circumvent this issue, we establish a relationship between the score \emph{differences} across different pairs.
For this purpose, we define $s_X$, $s^m_X$, and $\tilde s^m_X$ as the score function of the observed variable $X$ under $\mcE^0$, $\mcE^m$, and $\tilde \mcE^m$, respectively. Given any valid encoder $h$, based on \eqref{eq:encoder}, the estimated latent variable $\hat Z(X;h)$ and $X$ are related through $\hat Z(X;h)=h(X)$. We use this relationship to characterize the connection between score differences as follows, which is formalized in Lemma~\ref{lm:score-difference-transform-general} in Section~\ref{sec:analysis}.
\begin{align}
    &\mbox{between }\mcE^{0} \mbox{ and } \mcE^{m}:  \label{eq:score-difference-z-x-1} \\ 
    &\quad s_{\hat Z}(\hat z;h) - s^{m}_{\hat Z}(\hat z;h) = [J_{h^{-1}}(\hat z)]^{\top} \cdot [s_{X}(x) - s^{m}_{X}(x)] \ ,  \notag \\
    &\mbox{between }\mcE^{0} \mbox{ and } \tilde \mcE^{m}: \label{eq:score-difference-z-x-2}  \\
    &\quad s_{\hat Z}(\hat z;h) - \tilde s^{m}_{\hat Z}(\hat z;h) = [J_{h^{-1}}(\hat z)]^{\top} \cdot [s_{X}(x) - \tilde s^{m}_{X}(x)] \ , \notag \\
    &\mbox{between }\mcE^{m} \mbox{ and } \tilde \mcE^{m}: \label{eq:score-difference-z-x} \\
    &\quad s^{m}_{\hat Z}(\hat z;h) - \tilde s^{m}_{\hat Z}(\hat z;h) = [J_{h^{-1}}(\hat z)]^{\top} \cdot [s^{m}_{X}(x) - \tilde s^{m}_{X}(x)] \ .  \notag
\end{align}
We will show that among all valid encoders $h\in\mcH$, the true encoder $g^{-1}$ results in the minimum number of variations between the score estimates $s^{m}_{\hat Z}(\hat z;h)$ and $\tilde s^{m}_{\hat Z}(\hat z;h)$ (see Lemma~\ref{lm:min-score-variations}). To formalize these, corresponding to each valid encoder $h \in \mcH$ we define score change matrices $\bD_{\rm t}(h)$, $\bD(h)$, and $\tilde \bD(h)$ as follows. For all $i,m \in[n]$:
\begin{align}
    [\bD_{\rm t}(h)]_{i,m} \triangleq  \E\Big[\big|[s_{\hat Z}^{m}(\hat Z;h) - \tilde s_{\hat Z}^{m}(\hat Z;h)]_i\big|\Big] \ , \label{eq:Dh-entry} \\
    [\bD(h)]_{i,m} \triangleq \E\Big[\big|[s_{\hat Z}(\hat Z;h) - s_{\hat Z}^{m}(\hat Z;h)]_i\big|\Big] \ , \label{eq:Dh-obs-entry} \\
    [\tilde \bD(h)]_{i,m} \triangleq \E\Big[\big|[s_{\hat Z}(\hat Z;h) - \tilde s_{\hat Z}^{m}(\hat Z;h)]_i\big|\Big] \ , \label{eq:Dh-obs-tilde-entry}
\end{align}
where expectations are under the measures of latent score functions induced by the probability measure of observational data. The entry $[\bD_{\rm t}(h)]_{i,m}$ will be strictly positive only when there is a set of samples $X$ with a strictly positive measure that renders non-identical scores $s^{m}_{\hat Z}(\hat z;h)$ and $\tilde s^{m}_{\hat Z}(\hat z;h)$. Similar properties hold for the entries of $\bD(h)$ and $\tilde \bD(h)$ for the respective score functions. The GSCALE-I algorithm is summarized in Algorithm \ref{alg:main} and its key steps are described next.

\begin{algorithm}[t]
\caption{{\bf G}eneralized {\bf S}core-based {\bf Ca}usal {\bf L}atent {\bf E}stimation via {\bf I}nterventions (GSCALE-I)}
\label{alg:main}
\begin{algorithmic}[1]
\Require $\mcH$, samples of $X$ from environment $\mcE^0$ and environment sets $\mcE$ and $\tilde \mcE$, \texttt{is\_coupled}.
\Ensure Latent variable estimate $\hat Z$ and latent DAG estimate $\hat \mcG$.
\State \textbf{Step 1:} Compute score differences: $(s_X-s_X^m), (s_X-\tilde s_X^m)$, and $(s_X^m - \tilde s_X^m)$ for all $m\in[n]$.
\State \textbf{Step 2:} Identify the encoder by minimizing score variations:
\If{\texttt{is\_coupled}}
    \State Solve $\mcP_1$ in \eqref{eq:OPT1}, select a solution $h^*$.
\Else \Comment{search for the correct coupling}
    \ForAll{permutations $\pi$ of $[n]$}
        \State Temporarily relabel $\tilde \mcE^{m}$ to $\tilde \mcE^{\pi_m}$ for all $m\in[n]$, and solve $\mcP_2$ in \eqref{eq:OPT2}
        \State If there is a solution, select a solution $h^*$ and break from the loop.
    \EndFor
\EndIf
\State \textbf{Step 3:} Latent estimates: $\hat Z = h^*(X)$.
\State \textbf{Step 4:} Latent DAG recovery: Construct latent DAG $\hat \mcG$ using \eqref{eq:construct-graph}.
\State \Return $\hat Z$ and $\hat \mcG$.
\end{algorithmic}
\end{algorithm}%

\begin{description}[leftmargin=0em]
    \item[Inputs:] The inputs of GSCALE-I are the observed data from the observational and interventional environments, whether environments are coupled/uncoupled, and a set of valid encoders $\mcH$. 
    \item[Step 1 --  Score differences:] We start by computing score differences $(s_X-s_X^m), (s_X-\tilde s_X^m)$, and $(s_X^m - \tilde s_X^m)$ for all $m\in[n]$.  
    \item [Step 2 -- Identifying the encoder:] The key property in this step is that the number of variations of the estimated latent score differences is always no less than the number of variations of the true latent score differences. We have two different approaches for coupled and uncoupled settings.
    \begin{description}[leftmargin=1em]
    \item[Step 2 (a) -- Coupled environments:] We solve the following optimization problem
    \begin{equation}\label{eq:OPT1}
        \mcP_1\triangleq \left\{
        \begin{aligned}
            \min_{h\in\mcH}  \quad & \|\bD_{\rm t}(h)\|_0 \\
            \mbox{s.t.} \quad & \bD_{\rm t}(h) \;\; \mbox{is a diagonal matrix}\ .  
        \end{aligned}\right. 
    \end{equation} 
    Constraining $\bD_{\rm t}(h)$ to be diagonal enforces that the final estimate $\hat Z$ and $Z$ will be related by permutation $\mcI$ (the intervention order). We select a solution of $\mcP_1$ in \eqref{eq:OPT1} as our encoder estimate and denote it by $h^*$. \looseness=-1

    \item[Step 2 (b) -- Uncoupled environments:] In this setting, additionally, we need to determine the correct coupling between the interventional environment sets $\mcE$ and $\tilde \mcE$. To this end, we iterate through permutations $\pi$ of $[n]$, and temporarily relabel $\tilde \mcE^{m}$ to $\tilde \mcE^{\pi_m}$ for all $m \in [n]$ within each iteration. Subsequently, we solve the following optimization problem, 
    \begin{equation}\label{eq:OPT2}
       \hspace{-0.2in} \mcP_2\triangleq \left\{
        \begin{aligned}
            \min_{h\in\mcH}  \quad & \|\bD_{\rm t}(h)\|_0 \\
            \mbox{s.t.} \quad &  \bD_{\rm t}(h) \;\; \mbox{is a diagonal matrix}\\
            & \mathds{1}\{\bD(h)\}=\mathds{1}\{\tilde \bD(h)\} \\
                & \mathds{1}\{\bD(h)\} \odot \mathds{1}\{\bD^{\top}(h)\} = \bI_{n \times n} \ .
        \end{aligned}\right. 
    \end{equation}
    The constraint $\mathds{1}\{\bD(h)\}=\mathds{1}\{\tilde \bD(h)\}$ ensures that a permutation of the correct encoder is a solution to $\mcP_2$ if the coupling is correct, and the last constraint ensures that $\bD(h)$ does not contain 2-cycles. We will show that $\mcP_2$ is always feasible and, more specifically, admits a solution if and only if $\pi$ is the correct coupling (see Lemma~\ref{lm:opt2-solution}), in which case, we select a solution of $\mcP_2$ as our encoder estimate and denote it by $h^*$.    
    \end{description}

    \item[Step 3 -- Latent estimates:]
    The latent causal variables are estimated using $h^*$ via $\hat Z = h^*(X)$, where $X$ is the observational data. 
    
    \item[Step 4 -- Latent DAG recovery:] We construct DAG $\hat \mcG$ from $\bD(h^*)$ by assigning the non-zero coordinates of the $i$-th column of $\bD(h^*)$ as the parents of node $i$ in $\hat \mcG$, i.e.,
    \begin{equation} \label{eq:construct-graph}
        \hat{\Pa}(i) \triangleq \big\{j \neq i: [\bD(h^*)]_{j,i} \neq 0 \big\} \ , \quad \forall i \in [n] \ .
    \end{equation}
\end{description}
\begin{remark}\label{remark:achievability}
    For the nonparametric identifiability results, having an oracle that solves the functional optimization problems in~\eqref{eq:OPT1}~and~\eqref{eq:OPT2}, respectively, is sufficient. Solving these two problems in their most general form requires calculus of variations. These two problems, however, for any desired parameterized family of functions $\mcH$ (e.g., linear, polynomial, and neural networks), reduce to parametric optimization problems.
\end{remark}

\section{PROPERTIES OF GSCALE-I}\label{sec:analysis}

In this section, we analyze the properties and steps of GSCALE-I algorithm. The proofs of the results in this section are provided in Appendices~\ref{appendix:scores} and ~\ref{appendix:proofs-identifiability}. We start by presenting the key properties of score functions that play pivotal roles throughout the analysis.

\subsection{Properties of Score Functions}
We investigate the variations of the latent score functions that are caused by the atomic hard interventions. The following lemma delineates the set of coordinates of the score function that are affected under interventions in all relevant cases. 
\begin{lemma}[Score Changes]\label{lm:parent_change_comprehensive}
    \looseness=-1
    Consider environments $\mcE^0$, $\mcE^{m}$, and $\tilde \mcE^{m}$ with unknown intervention targets $I^m$ and $\tilde I^m$.
    \begin{enumerate}[label=(\roman*),leftmargin=2em, itemsep=-.05 in]
        \item Score functions $s$ and $s^{m}$ (or  $\tilde s^{m}$) differ in their $i$-th coordinate if and only if node $i$ or one of its children is intervened in $\mcE^{m}$ (or $\tilde \mcE^{m}$), i.e., 
        \begin{align}
            \label{eq:scorechanges_iff2}  \hspace{-0.15in} \E\Big[\big|[s(Z) - s^m(Z)]_i\big|\Big] \neq 0  &\iff i \in \overline{\Pa}(I^m)\ , \\
            \label{eq:scorechanges_iff3} \hspace{-0.15in} \E\Big[\big|[s(Z)] - \tilde s^m(Z)]_i\big|\Big] \neq 0 &\iff i \in \overline{\Pa}(\tilde I^m)\ . 
        \end{align}
    
        \item \textbf{Coupled environments $I^m = \tilde I^m$:} In the coupled environment setting, $s^{m}$ and $\tilde s^{m}$ differ in their $i$-th coordinate if and only if $i$ is intervened, i.e., 
         \begin{equation}\label{eq:scorechanges_iff4}
             \E\Big[\big|[s^m(Z) - \tilde s^m(Z)]_i\big|\Big] \neq 0 \iff i = I^m  \ . 
         \end{equation}
    
         \item \textbf{Uncoupled environments $I^m \neq \tilde I^m$:} Consider two interventional environments $\mcE^{m}$ and $\tilde \mcE^{m}$ with different intervention targets $I^m \neq \tilde I^m$. Consider additive noise models, in which 
         \begin{equation}\label{eq:additive}
             Z_i = f_i(Z_{\Pa(i)}) + N_i\ , 
         \end{equation}
         where functions $\{f_i : i \in [n]\}$ are general functions and $\{N_i : i \in [n]\}$ account for noise terms that have pdfs with full support. Given that $p$ is twice differentiable, the score functions $s^{m}$ and $\tilde s^{m}$ differ in their $i$-th coordinate if and only if node $i$ or one of its children is intervened,
         \begin{equation}\label{eq:scorechanges_iff5}
            \hspace{-0.38in}  \E\Big[\big|[s^{m}(Z)-\tilde s^{m}(Z)]_i\big|\Big] \neq 0 \iff i \in \overline{\Pa}(I^m,\tilde I^m)  \ .  
         \end{equation}
    \end{enumerate}
\end{lemma}

In the next lemma, we establish a transformation between the score differences across different environments for \emph{any} injective mapping $f$ from latent to observed space.
\begin{lemma}[Score Difference Transformation]\label{lm:score-difference-transform-general}
    Consider random vectors $Y_1,Y_2\in\R^r$ and $W_1$, $W_2 \in \R^s$ that are related through $Y_1=f(W_1)$ and $Y_2=f(W_2)$ such that $r \geq s$, probability measures of $W_1,W_2$ are absolutely continuous with respect to the $s$-dimensional Lebesgue measure, and $f: \R^s \to \R^r$ is an injective and continuously differentiable function. The difference of the score functions of $Y_1$ and $Y_2$, and that of $W_1$ and $W_2$ are related as
    \begin{equation}
        s_{W_1}(w)-s_{W_2}(w) = [J_f(w)]^{\top} \cdot [s_{Y_1}(y) - s_{Y_2}(y)] \ ,
    \end{equation}
    where $y=f(w)$ and $J_f(w) \in \R^s \to \R^r$ is the Jacobian of $f$ at point $w \in \R^s$.
\end{lemma}

We customize Lemma~\ref{lm:score-difference-transform-general} to two special cases. First, we consider score differences of $\hat Z(X;h)$ and $X$. By setting $f=h^{-1}$, Lemma~\ref{lm:score-difference-transform-general} immediately specifies the score differences of $\hat Z(X;h)$ and $X$ under different environment pairs in \eqref{eq:score-difference-z-x-1}, \eqref{eq:score-difference-z-x-2}, and \eqref{eq:score-difference-z-x}. Next, we consider score differences of $\hat Z(X;h)$ and $Z$. Note that $\hat Z(X;h) = h(X) = (h \circ g)(Z)$. Hence, by defining $\phi_h = h \circ g$ and setting $f=\phi_h^{-1}$, Lemma~\ref{lm:score-difference-transform-general} yields
\begin{align}
    &\mbox{between }\mcE^{0} \mbox{ and } \mcE^{m}: \label{eq:score-difference-z-zhat-1} \\ 
    &\quad s_{\hat Z}(\hat z;h) - s^{m}_{\hat Z}(\hat z;h) = [J_{\phi_h}(z)]^{-\top} \cdot [s(z) - s^{m}(z)] \ ,  \notag \\
    &\mbox{between }\mcE^{0} \mbox{ and } \tilde \mcE^{m}: \label{eq:score-difference-z-zhat-2}  \\
    &\quad s_{\hat Z}(\hat z;h) - \tilde s^{m}_{\hat Z}(\hat z;h) = [J_{\phi_h}(z)]^{-\top} \cdot [s(z) - \tilde s^{m}(z)] \ , \notag \\
    &\mbox{between }\mcE^{m} \mbox{ and } \tilde \mcE^{m}: \label{eq:score-difference-z-zhat} \\
    &\quad s^{m}_{\hat Z}(\hat z;h) - \tilde s^{m}_{\hat Z}(\hat z;h) = [J_{\phi_h}(z)]^{-\top} \cdot [s^{m}(z) - \tilde s^{m}(z)] \ .  \notag
\end{align}
Equipped with these results, we analyze the main algorithm steps.

\subsection{Analysis of Algorithm Steps}
The key idea for identifying the true encoder is that the number of variances between the score estimates $s^{m}_{\hat Z}(\hat z;h)$ and $\tilde s^{m}_{\hat Z}(\hat z;h)$ is minimized under the true encoder $g^{-1}$. To show that, first, we define the \emph{true} score change matrix $\bD_{\rm t}$ with entries for all $i,m \in [n]$,
\begin{equation}\label{eq:true-score-change-matrix}
    [\bD_{\rm t}]_{i,m} \triangleq \E\Big[\big|[s^m(Z) - \tilde s^m(Z)]_i\big|\Big] \ .
\end{equation} 
We start by considering coupled environments. Since the only varying causal mechanism across $\mcE^{m}$ and $\tilde\mcE^{m}$ is the intervened node $I^m=\tilde I^m$, based on \eqref{eq:scorechanges_iff3}, 
\begin{equation}
    \E\Big[\big|[s^m(Z)-\tilde s^m(Z)]_i \big|\Big] \neq 0  \iff i = I^m   \ ,
\end{equation}
which implies that $\mathds{1}\{\bD_{\rm t}\}$ is a permutation matrix. Specifically, $\mathds{1}\{\bD_{\rm t}\} = \bP_{\mcI}^{\top}$. We show that the number of variations between the score estimates $s_{\hat Z}^m(\hat z;h)$ and $\tilde s_{\hat Z}^m(\hat z;h)$, i.e., $\ell_0$ norm of $\bD_{\rm t}(h)$, cannot be less than the number of variations under the true encoder $g^{-1}$, that is $n = \norm{\bD_{\rm t}}_0$.
\begin{lemma}[Score Change Matrix Density]\label{lm:min-score-variations}
    For every $h \in \mcH$, the score change matrix $\bD_{\rm t}(h)$ is at least as dense as the score change matrix $\bD_{\rm t}$ associated with the true latent variables, 
    \begin{equation}
        \norm{\bD_{\rm t}(h)}_0 \geq \norm{\bD_{\rm t}}_0 = n\ .
    \end{equation}
\end{lemma}

The rest of the proof of Theorem~\ref{th:two-hard-coupled} builds on Lemma~\ref{lm:min-score-variations} and shows that, for any solution $h^*$ to \eqref{eq:OPT1}, we have $\mathds{1}\{J_{\phi_{h^*}}^{-1}(z)\} = \bP_{\mcI}^{\top}$ for all $z\in \R^n$. Finally, using Lemma~\ref{lm:parent_change_comprehensive}(i) we show that DAG $\hat \mcG$ constructed using $\bD(h^*)$ is isomorphic to the true latent DAG $\mcG$. We defer the complete proof to Appendix~\ref{appendix-proof-two-hard-coupled}.

Next, we consider the uncoupled environments. The proof consists of showing two properties of the optimization problem $\mcP_2$ specified in \eqref{eq:OPT2}: (i) it does not have a feasible solution if the coupling is incorrect, and (ii) it has a feasible solution if the coupling is correct, which are given by following Lemma~\ref{lm:opt2-no-solution} and \ref{lm:opt2-solution}, respectively. 

\begin{lemma}[Feasibility]\label{lm:opt2-no-solution}
    If the coupling is incorrect, i.e., $\pi \neq \mcI$, the optimization problem in $\mcP_2$ in \eqref{eq:OPT2} does not have a feasible solution.
\end{lemma}

The main intuition in the proof of Lemma~\ref{lm:opt2-no-solution} is that the constraints of $\mcP_2$ cannot be satisfied simultaneously under an incorrect coupling. We prove it by contradiction. We assume that $h^*$ is a solution, hence, $\bD_{\rm t}(h^*)$ is diagonal and $\mathds{1}\{\bD(h)\}=\mathds{1}\{\tilde \bD(h)\}$. Then, by scrutinizing the \emph{eldest} mismatched node, we show that $\bD(h^*) \cdot \bD^{\top}(h^*)$ cannot be a diagonal matrix, which contradicts the premise that $h^*$ is a feasible solution. 
\begin{lemma}\label{lm:opt2-solution}
    If the coupling is correct, i.e., $\pi = \mcI$, $h=\pi^{-1} \circ g^{-1}$ is a solution to $\mcP_2$ in \eqref{eq:OPT2}, and yields $\norm{\bD_{\rm t}(h)}_0=n$.
\end{lemma}

Lemmas~\ref{lm:opt2-no-solution} and \ref{lm:opt2-solution} collectively prove Theorem~\ref{th:two-hard-uncoupled} identifiability as follows. We can search over the permutations of $[n]$ until $\mcP_2$ admits a solution $h^*$. By Lemma~\ref{lm:opt2-no-solution}, the existence of this solution means that coupling is correct. Note that when the coupling is correct, the constraint set of $\mcP_1$ is a subset of the constraints in $\mcP_2$. Furthermore, the minimum value of $\norm{\bD_{\rm t}(h)}_0$ is lower bounded by~$n$ (Lemma~\ref{lm:min-score-variations}), which is achieved by the solution $h^*$ (Lemma~\ref{lm:opt2-solution}). Hence, $h^*$ is also a solution to $\mcP_1$, and by Theorem~\ref{th:two-hard-coupled}, it satisfies perfect recovery of the latent DAG and the latent variables.

\section{EMPIRICAL EVALUATIONS}\label{sec:simulations}
We provide empirical assessments of the achievability guarantees. Specifically, we empirically evaluate the performance of the GSCALE-I algorithm for recovering the latent variables $Z$ and the latent DAG $\mcG$ under \emph{coupled} interventions on synthetic data by solving the optimization problem $\mcP_1$ in \eqref{eq:OPT1}. The evaluations pursue a two-fold purpose: (i)~evaluating the performance of GSCALE-I, and (ii)~showcasing settings for which the existing literature does not have an achievability result (i.e., a constructive algorithm) and provide only identifiability results for them. Hence, the achievability results in this section lack counterparts in the existing literature. We focus on a non-polynomial transform $g$ and a non-linear latent causal model.

\noindent\textbf{Data generation.}
To generate $\mcG$ we use the Erd{\H o}s-R{\' e}nyi model with density $0.5$ and $n\in\{5,8\}$ nodes. For the observational causal mechanisms, we adopt an additive noise model with
\begin{equation}
    Z_i =\sqrt{Z_{\Pa(i)}^{\top} \cdot \bA_{i} \cdot Z_{\Pa(i)}} + N_{i}\ , \label{eq:quadratic-model--general-experiments}
\end{equation}
where $\{\bA_{i}:i\in[n]\}$ are positive-definite matrices, and the noise terms are zero-mean Gaussian variables with variances $\sigma_{i}^2$ sampled randomly from ${\rm Unif}([0.5,1.5])$. For the two hard interventions on node $i$, $Z_i$ is set to $N_{q,i} \sim \mcN(0,\sigma_{q,i}^2)$ and $N_{\tilde q,i} \sim \mcN(0,\sigma_{\tilde q,i}^2)$. We set $\sigma_{q,i}^2 = \sigma_{i}^2+1$ and $\sigma_{\tilde q,i}^2 = \sigma_{i}^2+2$. We consider target dimension values $d\in\{5,8,25,100\}$. For each $(n,d)$ pair, we generate 100 latent graphs and $n_{\rm s}$ samples of $Z$ per graph, where we set $n_{\rm s}=100$ for $n=5$ and $n_{\rm s}=300$ for $n=8$. As the transformation, we consider a generalized linear model,
\begin{equation}\label{eq:transf_exp}
    X = g(Z) = \tanh (\bG \cdot Z) \ ,
\end{equation}
in which $\tanh$ is applied element-wise, and parameter $\mcG\in \R^{d \times n}$ is a randomly sampled full-rank matrix.

\noindent\textbf{Score functions.}
GSCALE-I computes the score differences $(s_X-s_X^m), (s_X-\tilde s_X^m)$, and $(s_X^m - \tilde s_X^m)$ for all $m\in[n]$ in Step~1. The design of GSCALE-I is agnostic to how Step~1 is performed, i.e., any reliable method for estimating these score differences can be adopted. On the other hand, we note that the \emph{perfect} identifiability guarantees formalized in Theorem~\ref{th:two-hard-coupled} rely on having \emph{perfect} score differences. In our experiments, we mainly adopt a score oracle that computes the score differences in Step~1 (see Appendix~\ref{appendix:simulations} for details). Unlike identifiability, for achievability, we need score estimates, which are inevitably noisy. For this purpose, we also adopt a score estimator, sliced score matching with variance reduction (SSM-VR) due to its efficiency and accuracy for downstream tasks~\citep{song2020sliced}.

\noindent\textbf{Candidate encoder and loss function.} Leveraging \eqref{eq:transf_exp}, we parameterize valid encoders as
\begin{equation}
    \hat Z = h(X) = \bH \cdot \arctanh (X) \ .
\end{equation}
To use gradient descent to learn parameters $\bH$ of $h$, we relax $\ell_0$ norm in \eqref{eq:OPT1} and instead minimize the element-wise $\ell_{1,1}$ norm $\norm{\bD_{\rm t}(h)}_{1,1}$ computed with empirical expectations. We also add proper regularization terms to ensure that the estimated parameter $\bH^*$ will be full-rank.

\begin{table}[t]
    \centering
        \caption{GSCALE-I for a quadratic causal model with  \textbf{two coupled hard} interventions.}
        \label{table:gscalei-quadratic}
        \begin{tabular}{cc|cc|cc}
            \toprule
              & & \multicolumn{2}{c|}{\bf perfect scores} & \multicolumn{2}{c}{\bf noisy scores} \\
             $n$ & $d$ & $  \ell(Z,\hat Z)$  & ${\rm SHD}(\mcG,\hat \mcG) $ &  $ \ell(Z,\hat Z)$ & ${\rm SHD}(\mcG,\hat \mcG) $ \\
            \midrule
            $5$ & $5$  & 0.03 & 0.12 & 1.19 & 5.1 \\
            $5$ & $25$ & 0.03 & 0.04 & 1.09 & 4.4 \\
            $5$ & $100$ & 0.04 & 0.02 & 0.86 & 5.0 \\
            \midrule
            $8$ & $8$ & 0.16 &  1.56 & 0.81 & 11.9 \\
            $8$ & $25$ & 0.20 & 1.55 & 0.69 & 10.5 \\
            $8$ & $100$ & 0.24 &  1.50 & 0.77  & 11.85  \\
            \bottomrule
        \end{tabular}
\end{table}
\noindent\textbf{Evaluation metrics.}
GSCALE-I ensures perfect latent and DAG recovery. For assessing the recovery of the latent DAG, we report structural Hamming distance (SHD) between $\hat \mcG$ and $\mcG$. For the recovery of latent variables, we report the normalized $\ell_2$ loss,  $\ell(Z,\hat Z)\triangleq\|Z-\hat Z\|_2 / \norm{Z}_2$.

\noindent\textbf{Observations.} Table~\ref{table:gscalei-quadratic} shows that by using true score differences $(s_X-s_X^m), (s_X-\tilde s_X^m)$, and $(s_X^m - \tilde s_X^m)$, we can almost perfectly recover the latent variables and the latent DAG for $n=5$ nodes. When we consider a larger graph with $n=8$ nodes, the normalized $\ell_2$ loss remains less than $0.25$. We note that $\mcG$ with $n=8$ nodes and density $0.5$ has an expected number of 14 edges. Hence, having an average SHD of approximately 1.5 edges indicates that GSCALE-I yields a high performance at recovering latent causal relationships even when the transformation estimate is reasonable but not perfect. We also observe that increasing dimension $d$ of the observational data does not degrade the performance, confirming our analysis that GSCALE-I is agnostic to the dimension of observations. Finally, Table~\ref{table:gscalei-quadratic} shows that GSCALE-I's performance degrades when using noisy scores computed via SSM-VR. This degradation is due to the errors introduced by imperfect score estimates. This performance gap can be improved by the advances in the approaches to score estimation.

\section{CONCLUSION}\label{sec:conclusion}

In this paper, we have established identifiability results for latent causal representations from two interventional environments per latent node without restrictions on transformation between latent and observed space, and the causal models. We addressed both \emph{uncoupled} and \emph{coupled} settings. Importantly, the learner does not which pair of environments share the same intervened node in the former setting, which improved upon the recent results in nonparametric CRL identifiability. Furthermore, we have shown that the faithfulness assumption that is required in previous studies in literature can be dispensed with given access to observational data. Finally, we provided an algorithm owing to the constructive proof technique based on variations in score functions under interventions and demonstrated its success at identifying both the inverse transformation and latent DAG via simulations on synthetic data.

One major direction for future work is relaxing the requirement of two atomic hard interventions per node. Partial identifiability guarantees for a non-exhaustive set of interventions can also be useful for making inferences from a reduced number of environments. Similarly, investigating the sufficient conditions for which a set of multi-target interventions guarantee identifiability is a promising research direction. Finally, comparing the experiments with perfect and noisy score estimates indicates that the practical bottleneck of our score-based framework is access to an accurate score difference estimator, which is an interesting future direction itself.

\subsubsection*{Acknowledgements}
The work of B. Var{\i}c{\i}, E. Acart{\"u}rk, and A. Tajer was supported by the Rensselaer-IBM AI Research Collaboration \href{http://airc.rpi.edu}{(http://airc.rpi.edu)}, part of the IBM AI Horizons Network \href{http://ibm.biz/AIHorizons}{(http://ibm.biz/AIHorizons)}.

\bibliographystyle{abbrvnat}
\bibliography{references}

\section*{Checklist}

 \begin{enumerate}

 \item For all models and algorithms presented, check if you include:
 \begin{enumerate}
   \item A clear description of the mathematical setting, assumptions, algorithm, and/or model. \textbf{Yes.}
   
   \item An analysis of the properties and complexity (time, space, sample size) of any algorithm. \textbf{Yes.} See Section~\ref{sec:analysis}.
   
   \item (Optional) Anonymized source code, with specification of all dependencies, including external libraries.  \textbf{Yes.} See the supplementary material. 
 \end{enumerate}

 \item For any theoretical claim, check if you include:
 \begin{enumerate}
   \item Statements of the full set of assumptions of all theoretical results. \textbf{Yes.}
   \item Complete proofs of all theoretical results. \textbf{Yes.} See Section~\ref{sec:analysis} and Appendix~\ref{appendix:proofs-identifiability}.
   \item Clear explanations of any assumptions. \textbf{Yes.}     
 \end{enumerate}

 \item For all figures and tables that present empirical results, check if you include:
 \begin{enumerate}
   \item The code, data, and instructions needed to reproduce the main experimental results (either in the supplemental material or as a URL). \textbf{Yes.} See Appendix~\ref{appendix:simulations} and supplementary material.
   \item All the training details (e.g., data splits, hyperparameters, how they were chosen). \textbf{Yes.}
         \item A clear definition of the specific measure or statistics and error bars (e.g., with respect to the random seed after running experiments multiple times). \textbf{Yes.}
         \item A description of the computing infrastructure used. (e.g., type of GPUs, internal cluster, or cloud provider). \textbf{Not Applicable.}
 \end{enumerate}

 \item If you are using existing assets (e.g., code, data, models) or curating/releasing new assets, check if you include:
 \begin{enumerate}
   \item Citations of the creator If your work uses existing assets. \textbf{Not Applicable.}
   \item The license information of the assets, if applicable. \textbf{Not Applicable.}
   \item New assets either in the supplemental material or as a URL, if applicable. \textbf{Not Applicable.}
   \item Information about consent from data providers/curators. \textbf{Not Applicable.}
   \item Discussion of sensible content if applicable, e.g., personally identifiable information or offensive content. \textbf{Not Applicable.}
 \end{enumerate}

 \item If you used crowdsourcing or conducted research with human subjects, check if you include:
 \begin{enumerate}
   \item The full text of instructions given to participants and screenshots. \textbf{Not Applicable.}
   \item Descriptions of potential participant risks, with links to Institutional Review Board (IRB) approvals if applicable. \textbf{Not Applicable.}
   \item The estimated hourly wage paid to participants and the total amount spent on participant compensation. \textbf{Not Applicable.}
 \end{enumerate}

 \end{enumerate}

\newpage
\onecolumn
\appendix

\def\toptitlebar{
\hrule height4pt
\vskip .25in}

\def\bottomtitlebar{
\vskip .25in
\hrule height1pt
\vskip .25in}

\hsize\textwidth
\linewidth\hsize \toptitlebar %
{\centering \Large \textbf{General Identifiability and Achievability for \\ Causal Representation Learning: \\ \hspace{2.2in} Supplementary Material} }

\bottomtitlebar 

\section{Related Work}\label{appendix:related}

\paragraph{Identifiable representation learning.}
One of the primary goals of representation learning is to identify the underlying latent factors responsible for generating observed data. However, as discussed in Section~\ref{sec:introduction}, latent factors are not identifiable unless there is auxiliary information or additional structure that explains the data generation process \citep{hyvarinen1999nonlinear, locatello2019challenging}. Various strategies have been developed to address this issue when there is no inherent causal relationship among the latent factors. Some notable approaches include incorporating posterior regularization \citep{kumar2020implicit}, leveraging knowledge about the mechanisms governing system dynamics \citep{ahuja2021properties}, and using weak supervision along with auxiliary information \citep{shu2019weakly}. Furthermore, non-linear independent component analysis (ICA) leverages side information in the form of structured time series to exploit temporal information \citep{hyvarinen2017nonlinear, halva2020hidden}, or knowledge of auxiliary variables that make latent variables conditionally independent \citep{pmlr-v108-khemakhem20a, khemakhem2020ice, hyvarinen2019nonlinear}. In a related context, the identifiability of deep generative models is studied without auxiliary information \citep{kivva2022identifiability}.

\paragraph{Causal representation learning.} In one approach to identifiability of CRL, several studies have investigated the setting when pairs of observations are available -- one before and one after a mechanism change (e.g., an intervention) for the same underlying realization of exogenous variables involved \citep{ahuja2022weakly,locatello2020weakly,vonkugelgen2021self,Yang_2021_CVPR,brehmer2022weakly}. A typical example of this setup is considering images of an object from different angles \citep{brehmer2022weakly}. From a causal perspective, these pairs can be regarded as counterfactual pairs. Alternatively, some studies use temporal sequences to identify causal variables in the presence of interventions \citep{lachapelle2022disentanglement, yao2022, lippe2022icitris}. However, we note that this paper does not focus on time-series data. Our approach operates under a milder form of supervision, where we can observe data under different interventional distributions in the latent space while the counterfactual instances remain unobservable.

\paragraph{CRL from interventions.} We repeat the related work discussed in Section~\ref{sec:introduction} here. The recent studies most closely related to the scope of this paper are~\citep{varici2023score}~and \citep{vonkugelgen2023twohard}. \citep{varici2023score} establishes an inherent connection between score functions and CRL, and based on that, designs a score-based CRL framework. Specifically, using \emph{one intervention per node} under a non-linear causal model and a linear transformation,~\citep{varici2023score} provides both identifiability and achievability results. It shows that finding the variations of the score functions across different intervention environments is sufficient to recover linear $g$ and $\mcG$ that have non-linear causal structures. We have three major distinctions from~\citep{varici2023score} in settings by assuming nonparametric choices of transformations $g$, a general latent causal model, and using two hard interventions. The study in~\citep{vonkugelgen2023twohard} considers nonparametric models for $g$ and the causal relationships and shows that two \emph{coupled} hard interventions per node suffice for identifiability. We have two major differences with~\citep{vonkugelgen2023twohard}. First, we assume uncoupled interventional environments, whereas~\citep{vonkugelgen2023twohard} focuses on coupled environments. Secondly, the approach of~\citep{vonkugelgen2023twohard} focuses mainly on identifiability (e.g., no algorithm for recovery of the latent variables), whereas we address both identifiability and achievability. 

The studies that focus on the \emph{parametric} settings include~\citep{liu2022identifying,squires2023linear,ahuja2023interventional,zhang2023identifiability,buchholz2023learning}. \citep{liu2022identifying} aims to learn latent causal graphs and identify latent representations. However, its focus is on linear Gaussian latent models, and its extensions to even non-linear Gaussian models are viable at the expense of restricting the graph structure. \citep{squires2023linear} considers linear causal models and proves identifiability under hard interventions and the impossibility of identifiability under soft interventions. This setting complements that of~\citep{varici2023score}, which considers non-linear latent causal models and proves identifiability under hard and milder identifiability guarantees under soft interventions. The study in \citep{ahuja2023interventional} considers a polynomial transform and shows that it can be reduced to an affine transform by an autoencoding process and proves identifiability under \emph{do} interventions or soft interventions on bounded latent variables. \citep{zhang2023identifiability} builds on the results of \citep{ahuja2023interventional}, considers polynomial transforms under non-linear causal models, and proves identifiability under soft interventions. Finally, \citep{buchholz2023learning} focuses on linear Gaussian causal models and extends the results of \citep{squires2023linear} to prove identifiability for general transforms. Other studies on the \emph{nonparametric} settings include~\citep{jiang2023learning,liang2023causal}. The study in~\citep{jiang2023learning} considers identifying the latent DAG without recovering latent variables, where it is shown that a restricted class of DAGs can be recovered. The study in~\citep{liang2023causal} assumes that the latent DAG is already known and recovers the latent variables under hard interventions.

\paragraph{Score functions in causality.}  The study in \citep{rolland2022score} uses score-matching to recover non-linear additive Gaussian noise models. The proposed method finds the topological order of causal variables but requires additional pruning to recover the full graph. \citep{montagna2023scalable} focuses on the same setting, recovers the full graph from Jacobian scores, and dispenses with the computationally expensive pruning stage. \citep{sanchez2023diffusion} uses score-matching to learn the topological order as well while significantly improving the training procedure. All of these studies are limited to observed causal variables, whereas in our case, we have a causal model in the latent space.

\section{Score Function Properties under Interventions}\label{appendix:scores}
In this section, we provide the proofs relating to score functions. First, we provide the following facts that will be used repeatedly in the proofs.
\begin{proposition}\label{prop:continuity-argument}
    Consider two continuous functions $f,g: \R^n \to \R$. Then, for any $\alpha > 0$,
    \begin{equation}
        \exists z \in \R^n \;\; f(z) \neq g(z) \quad \iff \quad \E\Big[\big|f(Z)-g(Z)\big|^{\alpha}\Big] \neq 0 \ .
    \end{equation}
    Specifically, for $\alpha = 1$, we have
    \begin{equation}
        \exists z \in \R^n \;\; f(z) \neq g(z) \quad \iff \quad \E\Big[\big|f(Z)-g(Z)\big|\Big] \neq 0 \ .
    \end{equation}
\end{proposition}
\proof If there exists $z \in \R^n$ such that $f(z)\neq g(z)$, then $f(z)-g(z)$ is non-zero over a non-zero-measure set due to continuity. Then, $\E[|f(Z)-g(Z)|^{\alpha}] \neq 0$ since $p$ (pdf of $Z$) has full support. On the other direction, if $f(z)=g(z)$ for all $z\in\R^n$, then $\E[|f(Z)-g(Z)|^{\alpha}]=0$. This means that $\E[|f(Z)-g(Z)|^{\alpha}]\neq 0$ implies that there exists $z\in\R^n$ such that $f(z) \neq g(z)$.

\subsection{Proof of Lemma~\ref{lm:parent_change_comprehensive}}\label{sec:proof-lm:parent_change}

\paragraph{Case (i)} The statement directly follows from Lemma 4 of \citep{varici2023score}.

\paragraph{Case (ii) Coupled environments.} Suppose that $I^m=\tilde I^m= \ell$. Following \eqref{eq:pz_m_factorized_hard} and \eqref{eq:pz_m_factorized_hard_tilde}, we have
\begin{align}
    s^{m}(z) &= \nabla_z \log q_{\ell}(z_{\ell}) +  \sum_{i \neq \ell} \nabla_z \log p_i(z_i \med z_{\Pa(i)}) \ , \label{eq:sz_m_decompose_hard_case_1} \\
    \mbox{and} \quad \tilde s^{m}(z) &= \nabla_z \log \tilde q_{\ell}(z_{\ell}) +  \sum_{i \neq \ell} \nabla_z \log p_i(z_i \med z_{\Pa(i)})\ . \label{eq:sz_m_decompose_hard_tilde_case_1}
\end{align}
Then, subtracting \eqref{eq:sz_m_decompose_hard_tilde_case_1} from \eqref{eq:sz_m_decompose_hard_case_1} and looking at $i$-th coordinate, we have
\begin{align}\label{eq:coupled-hard-sz-sz-tilde}
    \big[s^{m}(z) - \tilde s^{m}(z)\big]_i &= \dfrac{\partial \log q_{\ell}(z_{\ell})}{\partial z_i} -  \dfrac{\partial \log \tilde q_{\ell}(z_{\ell})}{\partial z_i} \ .
\end{align}
If $i\neq \ell$, the right-hand side is zero and we have $\big[s^{m}(z) - \tilde s^{m}(z)\big]_i=0$ for all $z$. On the other hand, if $i=\ell$, since $q_{\ell}(z_{\ell})$ and $\tilde q_{\ell}(z_{\ell})$ are distinct, there exists $z\in\R^n$ such that $q_{\ell}(z_{\ell})\neq \tilde q_{\ell}(z_{\ell})$. Subsequently, by Proposition~\ref{prop:continuity-argument}, we have $\E\big[\big|[s^m(Z)-\tilde s^m(Z)]_i \big|\big] \neq 0$.

\paragraph{Case (iii) Uncoupled environments.} 
Suppose that $I^m = \ell$ and $\tilde I^m= j$, and $\ell \neq j$. Following \eqref{eq:pz_m_factorized_hard}, we have 
\begin{align}
    s^{m}(z) &= \nabla_z \log q_{\ell}(z_{\ell}) +  \nabla_z \log p_j(z_j \med z_{\Pa(j)}) + \sum_{k \in [n] \setminus \{\ell,j\}} \nabla_z \log p_k(z_k \med z_{\Pa(k}) \ , \label{eq:sz_m_decompose_hard_proof} \\
    \mbox{and} \quad \tilde s^{m}(z) &= \nabla_z \log q_j(z_j) +  \nabla_z \log p_{\ell}(z_{\ell} \med z_{\Pa(\ell)}) + \sum_{k \in [n] \setminus \{\ell,j\}} \nabla_z \log p_k(z_k \med z_{\Pa(k)})\ . \label{eq:sz_m_decompose_hard_tilde_proof}
\end{align}
Then, subtracting \eqref{eq:sz_m_decompose_hard_tilde_proof} from \eqref{eq:sz_m_decompose_hard_proof} we have
\begin{align}
    s^{m}(z) - \tilde s^{m}(z) &= \;  \nabla_z \log q_{\ell}(z_{\ell}) +  \nabla_z \log p_j(z_j \med z_{\Pa(j)})  - \nabla_z \log q_j(z_j) -  \nabla_z \log p_{\ell}(z_{\ell} \med z_{\Pa(\ell)}) \ . \label{eq:score-difference-proof-1} 
\end{align}
Scrutinizing the $i$-th coordinate, we have
\begin{align}
    \big[s^{m}(z) - \tilde s^{m}(z)\big]_i &= \dfrac{\partial \log q_{\ell}(z_{\ell})}{\partial z_i} + \dfrac{\partial \log p_j(z_j \med z_{\Pa(j)}) }{\partial z_i} - \dfrac{\partial \log q_j(z_j)}{\partial z_i} - \dfrac{\partial \log p_{\ell}(z_{\ell} \med z_{\Pa(\ell)})}{\partial z_i}  \ . \label{eq:score-difference-proof-k}
\end{align}

\paragraph{Proof of $\E\Big[\big|[s^{m}(Z) - \tilde s^{m}(Z)\big]_i\big|\Big] \neq 0\ \implies i\in \overline{{\Pa}}(\ell,j)$:} Suppose that $i \notin \overline{\Pa}({\ell},j)$. Then, none of the terms in the RHS of \eqref{eq:score-difference-proof-k} is a function of $z_i$. Therefore, all the terms in the RHS of \eqref{eq:score-difference-proof-k} are zero, and we have $\big[s^{m}(z) - \tilde s^{m}(z)\big]_i = 0$ for all $z$. By Proposition~\ref{prop:continuity-argument}, $\E\big[\big|[s^{m}(Z)-\tilde s^{m}(Z)]_i\big|\big]=0$. This, equivalently, means that if $\E\big[\big|[s^{m}(Z)-\tilde s^{m}(Z)]_i\big|\big] \neq 0$, then $i \in \overline{\Pa}({\ell},j)$. 

\paragraph{Proof of $\E\Big[\big|[s^{m}(Z)-\tilde s^{m}(Z)]_i\big|\Big] \neq 0 \impliedby i\in \overline{{\Pa}}(\ell,j)$:} We prove it by contradiction. Assume that $\big[s^{m}(z) - \tilde s^{m}(z)\big]_i = 0$ for all $z$. Without loss of generality, let $\ell \notin \overline{\Pa}(j)$. 
\begin{description}[leftmargin=1em]
    \item[\quad If $i=\ell$.] In this case, \eqref{eq:score-difference-proof-k} is simplified to
    \begin{align}
         0 = \big[s^{m}(z) - \tilde s^{m}(z)\big]_{\ell} = \dfrac{\partial \log q_{\ell}(z_{\ell})}{\partial z_{\ell}} - \dfrac{\partial \log p_{\ell}(z_{\ell} \med z_{\Pa(\ell)})}{\partial z_{\ell}}  \ .  \label{eq:proof-k-equals-i}
    \end{align}
    If $\ell$ is a root node, i.e., $\Pa(\ell)=\emptyset$, \eqref{eq:proof-k-equals-i} implies that $(\log q_{\ell})' (z_{\ell}) = (\log p_{\ell})'(z_{\ell})$ for all $z_{\ell}$. Integrating, we get $p_{\ell}(z_{\ell}) = \alpha q_{\ell}(z_{\ell})$ for some constant $\alpha$. Since both $p_{\ell}$ and $q_{\ell}$ are pdfs, they both integrate to one, implying $\alpha = 1$ and $p_{\ell}(z_{\ell}) = q_{\ell}(z_{\ell})$, which contradicts the premise that observational and interventional mechanisms are distinct. If $\ell$ is not a root node, consider some $k \in \Pa(\ell)$. Then, taking the derivative of \eqref{eq:proof-k-equals-i} with respect to $z_k$, we have
    \begin{align}\label{eq:proof-pi-by-zi-zl}
        0 = \dfrac{\partial^2 \log p_{\ell}(z_{\ell} \med z_{\Pa(\ell)})}{\partial z_{\ell} \partial z_k} \ . 
    \end{align}
    Recall the equation $Z_{\ell} = f_{\ell}(Z_{\Pa(\ell)}) + N_{\ell}$ for additive noise models specified in \eqref{eq:additive}. Denote the pdf of the noise term $N_{\ell}$ by $p_{N_\ell}$. Then, the conditional pdf $p_{\ell}(z_{\ell} \med z_{\Pa(\ell)})$ is given by $p_{\ell}(z_{\ell} \med z_{\Pa(\ell)}) = p_{N}(z_{\ell} - f_{\ell}(z_{\Pa(\ell)}))$. Denoting the score function of $p_{N_\ell}$ by $r_{\ell}$,
    \begin{align}\label{eq:score-of-additive-noise}
        r_{\ell}(u) \triangleq \dfrac{\rm d}{{\rm d} u} \log p_{N}(u) \ ,
    \end{align}
    we have
    \begin{align}\label{eq:additive-component-score}
        \dfrac{\partial \log p_{\ell}(z_{\ell} \med z_{\Pa(\ell)})}{\partial z_{\ell}} =  \dfrac{\partial \log p_{N}(z_{\ell} - f_{\ell}(z_{\Pa(\ell)}))}{\partial z_{\ell}} = r_{\ell}(z_{\ell} - f_{\ell}(z_{\Pa(\ell)})) \ .
    \end{align}
    Substituting this into \eqref{eq:proof-pi-by-zi-zl}, we obtain
    \begin{align}
        0 &= \dfrac{\partial r_{\ell}\big(z_{\ell} - f_{\ell}(z_{\Pa(\ell)})\big)}{\partial z_k} = - \dfrac{\partial f_{\ell}(z_{\Pa(\ell)})}{\partial z_k} \cdot r_{\ell}'\big(z_{\ell}-f_{\ell}(z_{\Pa(\ell)})\big) \ , \quad \forall z \in \R^n \ . \label{eq:proof-pi-by-zizl-additive}
    \end{align}
    Since $k$ is a parent of $\ell$, there exists a fixed $Z_{\Pa(\ell)} = z_{\Pa(\ell)}^*$ realization for which $\partial f_{\ell}(z_{\Pa(\ell)}^*)/\partial z_k$ is non-zero. Otherwise, $f_{\ell}(z_{\Pa(\ell)})$ would not be sensitive to $z_k$ which is contradictory to $k$ being a parent of $\ell$. Note that $Z_{\ell}$ can vary freely after fixing $Z_{\Pa(\ell)}$. Therefore, for \eqref{eq:proof-pi-by-zizl-additive} to hold, the derivative of $r_{p,\ell}$ must always be zero. However, the score function of a valid pdf with full support cannot be constant. Therefore, $\big[s^{m}(z)_i - \tilde s^{m}(z)\big]_i$ is not always zero, and we have $\E\big[\big|[s^{m}(Z) - \tilde s^{m}(Z)]_i\big|\big] \neq 0$.   

    \item[\quad If $i\neq \ell$]. In this case, \eqref{eq:score-difference-proof-k} is simplified to
    \begin{align}
         0 = \big[s^{m}(z) - \tilde s^{m}(z)\big]_i = \dfrac{\partial \log p_j(z_j \med z_{\Pa(j)}) }{\partial z_i} - \dfrac{\partial \log q_j(z_j)}{\partial z_i} - \dfrac{\partial \log p_{\ell}(z_{\ell} \med z_{\Pa(\ell)})}{\partial z_i}  \ . \label{eq:proof-k-notequals-i-1}
    \end{align}
    We investigate case by case and reach a contradiction for each case. First, suppose that $i \notin \Pa({\ell})$. Then, we have $i \in \overline{\Pa}(j)$, and \eqref{eq:proof-k-notequals-i-1} becomes
    \begin{align}
         0 = \big[s^{m}(z) - [\tilde s^{m}(z)\big]_i = \dfrac{\partial \log p_j(z_j \med z_{\Pa(j)}) }{\partial z_i} - \dfrac{\partial \log q_j(z_j)}{\partial z_i}  \ . \label{eq:proof-k-notequals-i-2} 
    \end{align}
    If $i = j$, the impossibility of \eqref{eq:proof-k-notequals-i-2} directly follows from the impossibility of \eqref{eq:proof-k-equals-i}. The remaining case is $i \in \Pa(j)$. In this case, taking the derivative of the right-hand side of \eqref{eq:proof-k-notequals-i-2} with respect to $z_j$, we obtain
    \begin{align}
           0 = \dfrac{\partial^2 \log p_j(z_j \med z_{\Pa(j)})}{\partial z_i \partial z_j} \ ,
    \end{align}    
    which is a realization of \eqref{eq:proof-pi-by-zi-zl} for $i \in \Pa(j)$ and $j$ in place of $k \in \Pa(\ell)$ and $\ell$, which we proved to be impossible in $i=\ell$ case. Therefore, $i \notin \Pa(\ell)$ is not viable. Finally, suppose that $i \in \Pa({\ell})$. Then, taking the derivative of the right-hand side of \eqref{eq:proof-k-notequals-i-1} with respect to $z_{\ell}$, we obtain 
    \begin{align}\label{eq:proof-pi-by-zi-zk}
           0 = \dfrac{\partial^2 \log p_{\ell}(z_{\ell} \med z_{\Pa(\ell)})}{\partial z_i \partial z_{\ell}} \ ,
    \end{align}
    which is again a realization of \eqref{eq:proof-pi-by-zi-zl} for $k = i$, which we proved to be impossible.
\end{description}
Hence, we showed that $\big[s^{m}(z) - \tilde s^{m}(z)\big]_i$ cannot be zero for all $z$ values. Then, by Proposition~\ref{prop:continuity-argument} we have $\E\big[\big|[s^{m}(Z) - \tilde s^{m}(Z)]_i\big|\big] \neq 0$, and the proof is concluded.

\subsection{Proof of Lemma \ref{lm:score-difference-transform-general}}\label{sec:proof-lm:score-difference}

Let us recall the setting. Consider random vectors $Y_1,Y_2\in\R^r$ and $W_1$, $W_2 \in \R^s$ that are related through $Y_1=f(W_1)$ and $Y_2=f(W_2)$ such that $r \geq s$, probability measures of $W_1,W_2$ are absolutely continuous with respect to the $s$-dimensional Lebesgue measure and $f: \R^s \to \R^r$ is an injective and continuously differentiable function.

In this setting, the realizations of $W_1$ and $Y_1$, and that of $W_2$ and $Y_2$, are related through $y = f(w)$. Since $f$ is injective and continuously differentiable, volume element ${\rm d}w$ in $\R^{s}$ gets mapped to $\left|\det([J_{f}(w)]^{\top} \cdot J_{f}(w))\right|^{1/2}\ {\rm d}w$ on $\image(f)$. Since $W_1$ has density $p_{W_1}$ absolutely continuous with respect to the $s$-dimensional Lebesgue measure, using the area formula~\citep{boothby2003introduction}, we can define a density for $Y_1$, denoted by $p_{Y_1}$, supported only on manifold $\mcM \triangleq \image(f)$ which is absolutely continuous with respect to the $s$-dimensional Hausdorff measure:
\begin{equation}
    p_{Y_1}(y) = p_{W_1}(w) \cdot \left|\det([J_{f}(w)]^{\top} \cdot J_{f}(w))\right|^{-1/2} \ , \quad \mbox{where} \quad y = f(w) \ . \label{eq:px-py-via-g-jacobian}
\end{equation}
Densities $p_{Y_2}$ and $p_{W_2}$ of $Y_2$ and $W_2$ are related similarly. Subsequently, score functions of $\{W_1,W_2\}$ and $\{Y_1,Y_2\}$ are specified similarly to \eqref{eq:sz-def} and \eqref{eq:sx-def}, respectively. Denote the Jacobian matrix of $f$ at point $w \in \R^{s}$ by $J_{f}(w)$, which is an $r \times s$ matrix with entries given by
\begin{equation}
    \big[J_{f}(w)\big]_{i,j} = \pdv{\big[f(w)\big]_i}{w_j}(x) = \pdv{y_i}{w_j} \ , \quad \forall i \in [r] \; , j \in [s] \ . \label{eq:g-jacobian-defn}
\end{equation}
Next, consider a function $\phi \colon \mcM \to \R$. Since the domain of $\phi$ is a manifold, its differential, denoted by $D\phi$, is defined according to \eqref{eq:Df-defn}. By noting $y = f(w)$, we can also differentiate $\phi$ with respect to $w \in \R^{s}$ as \citep[p.~57]{simon2014introduction}
\begin{equation}
    \nabla_{w} \phi(y) = \nabla_{w} (\phi \circ f)(w) = \big[J_{f}(w)\big]^{\top} \cdot D \phi(y) \ . \label{eq:nabla-x-fy}
\end{equation}
Next, given the identities in \eqref{eq:px-py-via-g-jacobian} and \eqref{eq:nabla-x-fy}, we find the relationship between score functions of $W_1$ and $Y_1$ as follows.
\begin{align}
    s_{W_1}(w)
    &= \nabla_{w} \log p_{W_1}(w) \label{eq:score-transform-proof-tmp1} \\
    \overset{\eqref{eq:px-py-via-g-jacobian}}&{=} \nabla_{w} \log p_{Y_1}(y) + \nabla_{w} \log \left|\det([J_{f}(w)]^{\top} \cdot J_{g}(w))\right|^{1/2}  \\
    \overset{\eqref{eq:nabla-x-fy}}&{=} \big[J_{f}(w)\big]^{\top} \cdot D \log p_{Y_1}(y) + \nabla_{w} \log \left|\det([J_{f}(w)]^{\top} \cdot J_{f}(w))\right|^{1/2} \\
    &= \big[J_{f}(w)\big]^{\top} \cdot s_{Y_1}(y) + \nabla_{w} \log \left|\det([J_{f}(w)]^{\top} \cdot J_{f}(w))\right|^{1/2} \ . \label{eq:sx-sz-for-g}
\end{align}
Following the similar steps that led to \eqref{eq:sx-sz-for-g} for $W_2$ and $Y_2$, we obtain
\begin{align}
    s_{W_2}(w) &= \big[J_{f}(w)\big]^{\top} \cdot s_{Y_2}(y) + \nabla_{w} \log \left|\det([J_{f}(w)]^{\top} \cdot J_{f}(w))\right|^{1/2} \ . \label{eq:sxm-szm-for-g} 
\end{align}
Subtracting \eqref{eq:sxm-szm-for-g} from \eqref{eq:sx-sz-for-g}, we obtain the desired result
\begin{align}
    s_{W_1}(w)-s_{W_2}(w) = \big[J_f(w)\big]^{\top} \cdot \big[s_{Y_1}(y) - s_{Y_2}(y)\big] \ . \label{eq:sw-from-sy-in-proof}
\end{align}

\begin{corollary}\label{corollary:sx-from-sz}
  Under the same setting and the assumptions as Lemma~\ref{lm:score-difference-transform-general}, we have
    \begin{equation}
        s_{Y_1}(y) - s_{Y_2}(y) = \Big[\big[J_f(w)\big]^{\dag}\Big]^{\top} \cdot \big[s_{W_1}(w)-s_{W_2}(w)\big] \ , \quad \mbox{where} \;\; y=f(w) \ .
    \end{equation}
\end{corollary}
\proof Multiplying \eqref{eq:sw-from-sy-in-proof} from left with $\big[[J_f(w)]^{\dag}\big]^{\top}$, we obtain
\begin{equation}
    \Big[\big[J_f(w)\big]^{\dag}\Big]^{\top} \cdot \big[s_{W_1}(w)-s_{W_2}(w)\big] = \Big[\big[J_f(w)\big]^{\dag}\Big]^{\top} \cdot \big[J_f(w)\big]^{\top} \cdot \big[s_{Y_1}(y) - s_{Y_2}(y)\big] \ . \label{eq:jac-rel-left-multiplied}
\end{equation}
Note that
\begin{equation}
    \Big[\big[J_f(w)\big]^{\dag}\Big]^{\top} \cdot \big[J_f(w)\big]^{\top} = J_f(w) \cdot \big[J_f(w)\big]^{\dag} \ . \label{eq:jjw-col-proj-eq}
\end{equation}
Note that, by properties of the Moore-Penrose inverse, for any matrix $\bA$, we have $\bA \cdot \bA^{\dag} \cdot \bA = \bA$. This means that $\bA \cdot \bA^{\dag}$ acts as a left identity for vectors in the column space of $\bA$. By definition, $s_{Y_1}$ and $s_{Y_2}$ have values in $T_{w} \image(f)$, the tangent space of the image manifold $f$ at point $w$. This space is equal to the column space of matrix $J_f(w)$. Therefore, $J_f(w) \cdot [J_f(w)]^{\dag}$ acts as a left identity for $s_{Y_1}(y)$ and $s_{Y_2}(y)$, and we have
\begin{equation}
    J_f(w) \cdot \big[J_f(w)\big]^{\dag} \cdot \big[s_{Y_1}(y) - s_{Y_2}(y)\big] = s_{Y_1}(y) - s_{Y_2}(y) \ . \label{eq:sy-in-col-jfw}
\end{equation}
Substituting \eqref{eq:jjw-col-proj-eq} and \eqref{eq:sy-in-col-jfw} into \eqref{eq:jac-rel-left-multiplied} completes the proof.

\section{Proofs of Identifiability Results}\label{appendix:proofs-identifiability}

In this section, we first prove identifiability in the coupled environments along with the observational environment case (Theorem~\ref{th:two-hard-coupled}). Then, we show that the result can be extended to coupled environments without observational environment (Theorem~\ref{th:faithfulness}) and uncoupled environments (Theorem~\ref{th:two-hard-uncoupled}).

For convenience, we recall the following equations from the main paper. For each $h\in\mcH$ we define $\phi_h \triangleq h \circ g$. Then, $\hat Z(X;h)$ and $Z$ are related as
\begin{align}
    \hat Z(X;h) =  h(X) = (h \circ g)(Z) = \phi_h (Z) \ .
\end{align}
Then, by setting $f=\phi_h^{-1}$, Lemma~\ref{lm:score-difference-transform-general} yields
\begin{align}
     \mbox{between }\mcE^{0} \mbox{ and } \mcE^m: \quad s_{\hat Z}(\hat z;h) - s^{m}_{\hat Z}(\hat z;h) &= \big[J_{\phi_h}(z)\big]^{-\top} \cdot \big[s(z) - s^{m}(z)\big] \ , \label{eq:score-difference-z-zhat-1-proof} \\
 \mbox{between }\mcE^{0} \mbox{ and } \tilde\mcE^m: \quad   s_{\hat Z}(\hat z;h) - \tilde s^{m}_{\hat Z}(\hat z;h) &= \big[J_{\phi_h}(z)\big]^{-\top} \cdot \big[s(z) - \tilde s^{m}(z)\big] \ ,  \label{eq:score-difference-z-zhat-2-proof} \\
   \mbox{between }\mcE^m \mbox{ and } \tilde\mcE^m: \quad s^{m}_{\hat Z}(\hat z;h) - \tilde s^{m}_{\hat Z}(\hat z;h) &= \big[J_{\phi_h}(z)\big]^{-\top} \cdot \big[s^{m}(z) - \tilde s^{m}(z)\big] \ . \label{eq:score-difference-z-zhat-proof}
\end{align}

\subsection{Proof of Theorem~\ref{th:two-hard-coupled}}\label{appendix-proof-two-hard-coupled}

First, we investigate the perfect recovery of latent variables.    
\paragraph{Recovering the latent variables.}
We recover the latent variables using only the coupled interventional environments $\{(\mcE^m,\tilde \mcE^m): m \in [n]\}$. Let $\rho$ be the permutation that maps $\{1,\dots,n\}$ to $\mcI$, i.e., $I^{\rho_i}=i$ for all $i\in[n]$ and $\bP_{\rho}$ to denote the permutation matrix that corresponds to $\rho$, i.e.,
\begin{align}\label{eq:intervention-order}
    [\bP_{\rho}]_{i,m} = \begin{cases}
        1  \ , &m = \rho_i \ , \\
        0 \ , & \text{else} \ .
    \end{cases}
\end{align}
Since we consider coupled atomic interventions, the only varying causal mechanism between $\mcE^{\rho_i}$ and $\tilde\mcE^{\rho_i}$ is that of the intervened node in $I^{\rho_i}=\tilde I^{\rho_i} = i$. Then, by Lemma~\ref{lm:parent_change_comprehensive}(ii), we have
\begin{align}\label{eq:diagonal-score-change}
    \E\bigg[\Big|\big[s^m(Z)- \tilde s^m(Z)\big]_k \Big|\bigg] \neq 0 \quad \iff \quad k = i  \ .
\end{align}
We recall the definition of true score change matrix $\bD_{\rm t}$ in \eqref{eq:true-score-change-matrix} with entries for all $i,m \in [n]$,
\begin{align}
    \big[\bD_{\rm t}\big]_{i,m} \triangleq \E\bigg[\Big|\big[s^m(Z)-\tilde s^m(Z)\big]_i\Big|\bigg] \ .
\end{align}
Then, using \eqref{eq:diagonal-score-change}, we have
\begin{equation}
    \mathds{1}\{\bD_{\rm t}\} = \bP_{\rho} = \bP_{\mcI}^{\top} \ .
\end{equation}
Next, we show that the number of variations between the score estimates $s_{\hat Z}^m(\hat z;h)$ and $\tilde s_{\hat Z}^m(\hat z;h)$ cannot be less than the number of variations under the true encoder $g^{-1}$, that is $n = \norm{\bD_{\rm t}}_0$. First, we provide the following linear algebraic property.

\begin{proposition}\label{prop:permutation-diagonal}
    If $\bA \in \R^{n \times n}$ is a full-rank matrix, then there exists a permutation matrix $\bP$ such that the diagonal elements of $\, \bP \cdot \bA \, $ are non-zero. 
\end{proposition}
\noindent\textbf{Proof } See Appendix~\ref{proof:permutation-diagonal}.

\renewcommand{\thelemma}{\ref{lm:min-score-variations}}
\begin{lemma}[Score Change Matrix Density]
    For every $h \in \mcH$, the score change matrix $\bD_{\rm t}(h)$ is at least as dense as the score change matrix $\bD_{\rm t}$ associated with the true latent variables, 
    \begin{align}
        \norm{\bD_{\rm t}(h)}_0 \geq \norm{\bD_{\rm t}}_0 = n\ .
    \end{align}
\end{lemma}
\renewcommand{\thelemma}{\arabic{lemma}}

\begin{proof}
Recall the definition of score change matrix $\bD_{\rm t}(h)$ in \eqref{eq:Dh-entry}. Using \eqref{eq:score-difference-z-zhat-proof}, we can write entries of $\bD_{\rm t}(h)$ equivalently as 
\begin{align}
    \big[\bD_{\rm t}(h)\big]_{i,m} =  \E\bigg[\Big|\big[J_{\phi_h}^{-\top}(Z)\big]_i \cdot \big[s^{m}(Z)-\tilde s^{m}(Z)\big]\Big|\bigg] \ , \quad \forall \; i,m \in [n] \ . \label{eq:Dh-entry-z}
\end{align}
Since $\phi_h = h \circ g$ is a diffeomorphism, $\big[J_{\phi_h}^{-\top}(z)\big]$ is full-rank for all $(h,z) \in \mcH \times \R^n$. Using Proposition~\ref{prop:permutation-diagonal}, for all $(h,z)$, there exists a permutation $\pi(h,z)$ of $[n]$ with permutation matrix 
$\bP_1(h,z)$ such that 
\begin{align}
    \Big[\bP_1(h,z) \cdot J_{\phi_h}^{-\top}(z) \Big]_{i,i} \neq 0 \ , \quad \forall i \in [n] \ , \quad \mbox{where} \quad \bP_1(h,z) \triangleq \bP_{\pi(h,z)} \ .
\end{align}
Next, recall that interventional discrepancy means that, for each $i\in [n]$, there exists a null set $\mcT_i \subset \R$ such that $[s^{\rho_i}(z)]_i \neq [\tilde s^{\rho_i}(z)]_i$ for all $z_i \in \\mcK \setminus \mcT_i$ (regardless of the value of other coordinates of $z$). Then, there exists a null set $\mcT \subset \R^n$ such that $[s^{\rho_i}(z)]_i \neq [\tilde s^{\rho_i}(z)]_i$ for all $i\in[n]$ for all $z\in \R^n \setminus \mcT$. We denote this set $\R^n \setminus \mcT$ by $\mcZ$ as follows:
\begin{align}\label{eq:set-all-varying-score-points}
    \mcZ \triangleq \{ z \in \R^n  : \;\; \big[s^{\rho_i}(z)\big]_i \neq \big[\tilde s^{\rho_i}(z)\big]_i \ , \quad \forall i \in [n]  \} \ .
\end{align}
Then, for all $z \in \mcZ$, $h \in \mcH$, and $i \in [n]$, we have
\begin{align}
    \big[\bD_{\rm t}(h)\big]_{\pi_i(h,z),\rho_i} &= \E\bigg[\Big|\big[J_{\phi_h}^{-\top}(Z)\big]_{\pi_i(h,z)} \cdot \big[s^{\rho_i}(Z)-\tilde s^{\rho_i}(Z)\big]\Big|\bigg] \\
     &= \E\bigg[\Big|\big[J_{\phi_h}^{-\top}(Z)]_{\pi_i(h,z),i} \cdot \big[s^{\rho_i}(Z)_i - \tilde s^{\rho_i}(Z)\big]_i\Big|\bigg] \ ,
\end{align}
in which $\pi_i(h,z)$ denotes the $i$-th element of the permutation $\pi(h,z)$. By the definition of $\pi(h,z)$, for any $z \in \mcZ$, we know that $\big[J_{\phi_h}^{-\top}(z)\big]_{\pi_i(h,z),i} \neq 0$. Furthermore, by the definition of $\mcZ$, we have $\big[s^{\rho_i}(z) - \tilde s^{\rho_i}(z)\big]_i \neq 0$ for $z\in \mcZ$. Then, we have $\big[\bD_{\rm t}(h)\big]_{\pi_i(h,z),\rho_i} \neq 0$, which implies
\begin{align}
    \mathds{1}\big\{\bD_{\rm t}(h)\big\} \succcurlyeq \bP_1^{\top}(h,z) \cdot \bP_{\rho} \ , \quad \forall h \in \mcH, \;\; \forall z \in \mcZ \ . \label{eq:Du-lower-bound} 
\end{align}
Therefore, $\norm{\bD_{\rm t}(h)}_0 \geq \norm{\bP_{\rho}}_0 = n$ for any $h \in \mcH$, and the proof is concluded since we have $\mathds{1}\{\bD_{\rm t}\}=\bP_{\rho}$. 
\end{proof}

The lower bound for $\ell_0$ norm is achieved if and only if $\mathds{1}\{\bD_{\rm t}(h)\}=\bP_{\rho}$, which is an unknown permutation matrix. Since the only diagonal permutation matrix is $\bI_{n \times n}$, the solution set of the constrained optimization problem in \eqref{eq:OPT1} is given by
\begin{align}\label{eq:refine-candidates-two-hard}
    \mcH_1 \triangleq \big\{h \in \mcH: \mathds{1}\{\bD_{\rm t}(h)\} = \bI_{n \times n} \big\} \ .
\end{align}
Next, consider some fixed solution $h^* \in \mcH_1$. Due to \eqref{eq:Du-lower-bound}, we have
\begin{equation}
    \mathds{1}\{\bD_{\rm t}(h^*)\} =\bI_{n \times n} \succcurlyeq \bP_1^{\top}(h^*,z) \cdot \bP_{\rho} \ ,
\end{equation}
which implies that we must have $\bP_1(h^*,z) = \bP_{\rho}$ for all $z \in \mcZ$. Then, $\pi_i(h^*,z)=\rho_i$ for all $i\in[n]$. We will show that for all $i\neq j$, we have
\begin{equation}\label{eq:jacobian-condition-coupled}
    \big[J_{\phi_{h^*}}^{-1}(z)\big]_{j,\rho_i} = 0 \ , \quad \forall z \in \R^n
\end{equation}
To show \eqref{eq:jacobian-condition-coupled}, first consider $i\neq j$, which implies $[\bD_{\rm t}(h^*)]_{\rho_i,\rho_j} = 0$ since $\mathds{1}\{\bD_{\rm t}(h^*)\}=\bI_{n \times n}$. Then, using \eqref{eq:Dh-entry-z}, $\mathds{1}\{\bD_{\rm t}(h^*)\}=\bI_{n \times n}$ and Lemma~\ref{lm:parent_change_comprehensive}(ii), we have
\begin{align}
     \big[\bD_{\rm t}(h^*)\big]_{\rho_i,\rho_j} &=  \E\bigg[\Big|\big[J_{\phi_{h^*}}^{-\top}(Z)\big]_{\rho_i} \cdot \big[s^{\rho_j}(Z)-\tilde s^{\rho_j}(Z)\big]\Big|\bigg] \\
     & =  \E\bigg[\Big|\big[J_{\phi_{h^*}}^{-1}(Z)\big]_{j,\rho_i} \cdot \big[s^{\rho_j}(Z) - \tilde s^{\rho_j}(Z)\big]_j\Big|\bigg] \\ 
     & = 0 \ .
\end{align}
Note that $[s^{\rho_j}(z) - \tilde s^{\rho_j}(z)]_j \neq 0$ for all $z\in \mcZ$. Hence, if $[J_{\phi_{h^*}}^{-1}(z)]_{j,\rho_i}$ is non-zero over a non-zero-measure set within $\mcZ$, then $[\bD_{\rm t}(h^*)]_{\rho_i,\rho_j}$ would not be zero. Therefore, $\big[J_{\phi_{h^*}}^{-1}(z)\big]_{j,\rho_i}=0$ on a set of measure $1$. Since $J_{\phi_{h^*}}^{-1}$ is a continuous function, this implies that 
\begin{equation}
    \big[J_{\phi_{h^*}}^{-1}(z)\big]_{j,\rho_i}=0 \ , \quad z \in \R^n \ .
\end{equation}
To see this, suppose that $[J_{\phi_{h^*}}^{-1}(z^*)]_{j,\rho_i} >0$ for some $z^* \in \mcZ$. Due to continuity, there exists an open set including $z^*$ for which $[J_{\phi_{h^*}}^{-1}(z^*)]_{j,\rho_i} >0$, and since open sets have non-zero measure, we reach a contradiction. Therefore, if $i\neq j$, $[J_{\phi_{h^*}}^{-1}(z)]_{j,\rho_i}=0$ for all $z\in \R^n$. Since $J_{\phi_{h^*}}^{-1}(z)$ must be full-rank for all $z \in \R^n$, we have 
\begin{equation}\label{eq:jacobian-coupled-diagonal}
    [J_{\phi_{h^*}}^{-1}(z)]_{i,\rho_i}\neq 0 \ , \quad \forall z \in \R^n \ , \forall i \in [n] \ .
\end{equation}
Then, for any $h^* \in \mcH_1$, $[\hat Z(X;h^*)]_{\rho_i} = [\phi_{h^*}(Z)]_{\rho_i}$ is a function of only $Z_i$, and we have
\begin{align}\label{eq:diffeomorphism-rho}
    [\hat Z(X;h^*)]_{\rho_i} = \phi_{h^*}(Z_i) \ , \quad \forall i \in[n] \ , 
\end{align}
which concludes the proof.

\paragraph{Recovering the latent graph} Consider the selected solution $h^* \in \mcH_1$. We construct the graph $\hat \mcG$ as follows. We create $n$ nodes and assign the non-zero coordinates of $\rho_j$-th column of $\bD(h^*)$ as the parents of node $\rho_j$ in $\hat \mcG$, i.e., 
\begin{align}\label{eq:construct-G-Z-hat}
    \hat{\Pa}(\rho_j) &\triangleq \big\{\rho_i \neq \rho_j : [\bD(h^*)]_{\rho_i,\rho_j} \neq 0 \big\} \ , \quad \forall j \in [n] \ .
\end{align}
Using \eqref{eq:Dh-obs-entry} and \eqref{eq:score-difference-z-zhat-1-proof}, we have
\begin{align}
     \hat{\Pa}(\rho_j) \overset{\eqref{eq:Dh-obs-entry}}&{=} \left\{\rho_i \neq \rho_j : \E\bigg[\Big|\big[s_{\hat Z}(\hat Z;h^*) - s_{\hat Z}^{\rho_j}(\hat Z;h^*)\big]_{\rho_i}\Big|\bigg] \neq 0\right\}  \\
     \overset{\eqref{eq:score-difference-z-zhat-1-proof}}&{=} \left\{\rho_i \neq \rho_j : \E\bigg[\Big|\big[J_{\phi_{h^*}}^{-\top}(Z)\big]_{\rho_i} \cdot \big[s(Z)-\tilde s^{\rho_j}(Z)\big]\Big|\bigg] \neq 0 \right\} \\
     &= \left\{ \rho_i \neq \rho_j : \E\bigg[\Big|\big[J_{\phi_{h^*}}^{-\top}(Z)\big]_{\rho_i,i} \cdot \big[s(Z) - \tilde s^{\rho_j}(Z)\big]_i \Big|\bigg]  \neq 0 \right\} \ .
\end{align}
Since $[J_{\phi_{h^*}}^{-\top}(z)]_{\rho_i,i} \neq 0$ for all $z \in \R^n$, we have
\begin{align}
    \hat{\Pa}(\rho_j)  = \bigg\{\rho_i \neq \rho_j : \E\bigg[\Big|\big[s(Z) - \tilde s^{\rho_j}(Z)\big]_i \Big|\bigg] \neq 0 \bigg\} \ .
\end{align}
By Lemma~\ref{lm:parent_change_comprehensive}(i), $\E\big[\big|[s(Z)-\tilde s^{\rho_j}(Z)]_i \big|\big] \neq 0$ if and only if $i \in \overline{\Pa}(j)$. Therefore, \eqref{eq:construct-G-Z-hat} implies that $\rho_i \in \hat{\Pa}(\rho_j)$ if and only if $i \in \Pa(j)$, which shows that $\mcG$ and $\hat \mcG$ are related through a graph isomorphism by permutation $\rho$, which was defined as $\mcI^{-1}$.

\subsection{Proof of Theorem~\ref{th:faithfulness}} \label{appendix-proof-faithfulness}

In the proof of Theorem~\ref{th:two-hard-coupled}, we showed that coupled hard interventions (without using observational environment) are sufficient for recovering the latent variables. Then, in this proof, we just focus on recovering the latent graph. Specifically, we will show that if $p$ (pdf of $Z$) is adjacency-faithful to $\mcG$ and the latent causal model is an additive noise model, then we can recover $\mcG$ without having access to observational environment $\mcE^0$. By Lemma~\ref{lm:parent_change_comprehensive}(iii), true latent score changes across $\{\mcE^{\rho_i},\tilde \mcE^{\rho_j}\}$ gives us $\overline{\Pa}(i,j)$ for $i\neq j$. First, we use the perfect latent recovery result to show that Lemma~\ref{lm:parent_change_comprehensive}(iii) also applies to estimated latent score changes. Specifically, using $\eqref{eq:score-difference-z-zhat-proof}$ and $\mathds{1}\{J_{\phi_{h^*}}^{-1}\}=\bP_{\rho}$, we have
\begin{align}
    \big[s_{\hat Z}^{\rho_i}(\hat z;h^*) - \tilde s_{\hat Z}^{\rho_j}(\hat z;h^*)\big]_{\rho_k}
    &= \big[J_{\phi_{h^*}}^{-\top}(z)\big]_{\rho_k} \cdot \big[s^{\rho_i}(z) - \tilde s^{\rho_j}(z)\big] \\
    &=  \big[J_{\phi_{h^*}}^{-\top}(z)\big]_{\rho_k,k} \cdot \big[s^{\rho_i}(z) - \tilde s^{\rho_j}(z)\big]_k \ . 
\end{align}
Recall that we have found $\big[J_{\phi_{h^*}}^{-\top}(z)\big]_{\rho_k,k} \neq 0$ for all $z\in\R^n$ in \eqref{eq:jacobian-coupled-diagonal}. Then, we have
\begin{align}\label{eq:graph-iso-step-1}
    \E\bigg[\Big|\big[s_{\hat Z}^{\rho_i}(\hat Z;h^*) - \tilde s_{\hat Z}^{\rho_j}(\hat Z;h^*)\big]_{\rho_k}\Big|\bigg] \neq 0 \quad \iff \quad \E\bigg[\Big|\big[s^{\rho_i}(Z) - \tilde s^{\rho_j}(Z)\big]_k\Big|\bigg] \neq 0 \ .
\end{align}
Hence, by Lemma~\ref{lm:parent_change_comprehensive}(iii),
\begin{align}\label{eq:graph-iso-step-2}
    \E\bigg[\Big|\big[s_{\hat Z}^{\rho_i}(\hat Z;h^*) - \tilde s_{\hat Z}^{\rho_j}(\hat Z;h^*)\big]_{\rho_k}\Big|\bigg] \neq 0 \quad \iff \quad k \in \overline{\Pa}(i,j)\ .
\end{align}
Let us denote the graph $\mcG_{\rho}$ that is related to $\mcG$ by permutation $\rho$, i.e., $i\in\Pa(j)$ if and only if $\rho_i \in \Pa_{\rho}(\rho_j)$ for which $\Pa_{\rho}(\rho_j)$ denotes the parents of node $\rho_j$ in $\mcG_{\rho}$. Using \eqref{eq:graph-iso-step-2}, we have
\begin{align}
    \E\bigg[\Big|\big[s_{\hat Z}^{\rho_i}(\hat Z;h^*) - \tilde s_{\hat Z}^{\rho_j}(\hat Z;h^*)\big]_{\rho_k}\Big|\bigg] \neq 0 \quad \iff \quad  \rho_k \in \overline{\Pa}_{\rho}(\rho_i,\rho_j)\ .
\end{align}
In the rest of the proof, we will show how to obtain $\{\Pa_{\rho}(i) : i \in [n]\}$ using $\{\overline{\Pa}_{\rho}(i,j) : i,j \in [n], \; i \neq j\}$. Since $\mcG_{\rho}$ is a graph isomorphism of $\mcG$, this problem is equivalent to obtaining $\{\Pa(i) : i \in [n]\}$ using $\{\overline{\Pa}(i,j) : i,j \in [n], \; i \neq j\}$. Note that $\hat Z_i$ (which corresponds to node $i$ in $\mcG_{\rho}$) is intervened in environments $\mcE^i$ and $\tilde \mcE^i$. We denote the set of root nodes by
\begin{equation}
    \mcK \triangleq \{i\in[n] : \Pa(i) = \emptyset \} \ ,
\end{equation}
and also define
\begin{equation}
    \mcB_i \triangleq \, \bigcap_{j \neq i} \overline{\Pa}(i,j) \ , \quad \forall i \in [n] \ , \quad \mbox{and} \quad \mcB \triangleq \{i : |\mcB_i|=1\} \ .
\end{equation}
Note that $\overline{\Pa}(i) \subseteq \mcB_i$. Hence, $|\mcB_i|=1$ implies that $i$ is a root node. We investigate the graph recovery in three cases.
\begin{enumerate}[leftmargin=*]
    \item $|\mcB|\geq 3$: For any node $i\in[n]$, we have
    \begin{align}
    \overline{\Pa}(i) \subseteq \mcB_i \subseteq \bigcap_{j \in \mcK \setminus \{i\}} \overline{\Pa}(i,j) = \overline{\Pa}(i) \cup \Big\{\bigcap_{j \in \mcK \setminus \{i\}} \{ j \} \Big\} = \overline{\Pa}(i) \ .
    \end{align}
    Note that, the last equality is due to $\cap_{j \in \mcK \setminus \{i\}} \{ j \} = \emptyset$ since there are at least two root nodes excluding $i$. Then, $\mcB_i = \overline{\Pa}(i)$ for all $i \in [n]$ and we are done.
    
    \item $|\mcB|=2$: The two nodes in $\mcB$ are root nodes. If there were at least three root nodes, we would have at least three nodes in $\mcB$. Hence, the two nodes in $\mcB$ are the only root nodes. Subsequently, every $i \notin \mcB$ is also not in $\mcK$ and we have 
    \begin{align}
        \overline{\Pa}(i) \subseteq \mcB_i \subseteq \bigcap_{j \in \mcK} \overline{\Pa}(i,j) = \overline{\Pa}(i) \cup \Big\{\bigcap_{j \in \mcK}\{j\}\Big\} = \overline{\Pa}(i) \ .
    \end{align}
    Hence, $\mcB_i = \overline{\Pa}(i)$ for every non-root node $i$ and we already have the two root nodes in $\mcB$, which completes the graph recovery.
    
    \item $|\mcB|\leq 1$: First, consider all $(i,j)$ pairs such that $|\overline{\Pa}(i,j)|=2$. For such an $(i,j)$ pair, at least one of the nodes is a root node, otherwise $\overline{\Pa}(i,j)$ would contain a third node. Using these pairs, we identify all root nodes as follows. Note that a hard intervention on node $i$ makes $Z_i$ independent of all of its non-descendants, and all conditional independence relations are preserved under element-wise diffeomorphisms such as $\phi_{h^*}$. Then, using the adjacency-faithfulness assumption, we infer that
    \begin{itemize}
        \item if $\hat Z_i \ci \hat Z_j$ in $\mcE^i$ and $\hat Z_i \ci \hat Z_j$ in $\tilde \mcE^j$, then both $i$ and $j$ are root nodes.
        \item if $\hat Z_i \notci \hat Z_j$ in $\mcE^i$, then $i \rightarrow j$ and $i$ is a root node.
        \item if $\hat Z_i \notci \hat Z_j$ in $\tilde \mcE^j$, then $j \rightarrow i$ and $j$ is a root node.
    \end{itemize}
    This implies that by using at most two independence tests, we can determine whether $i$ and $j$ nodes are root nodes. Hence, by at most $n$ independence tests, we identify all root nodes. We also know that there are at most two root nodes. If we have two root nodes, then $\mcB_i = \overline{\Pa}(i)$ for all non-root nodes, and the graph is recovered. If we have only one root node $i$, then for any $j \neq i$ we have
    \begin{align}
        \overline{\Pa}(j) \subseteq \mcB_j \subseteq \overline{\Pa}(i,j) = \overline{\Pa}(j) \cup \{i\} \ .
    \end{align}
    Finally, if $\hat Z_j \ci \hat Z_i \mid \{\hat Z_\ell : \ell \in \mcB_j \setminus \{i\}\}$ in $\tilde \mcE^j$, we have $i \notin \overline{\Pa}(j)$ due to adjacency-faithfulness. Otherwise, we conclude that $i \in \overline{\Pa}(j)$. Hence, an additional $(n-1)$ conditional independence tests ensure the recovery of all $\overline{\Pa}(j)$ sets, and the graph recovery is complete. 
\end{enumerate}

\subsection{Proof of Lemma~\ref{lm:opt2-no-solution}}\label{proof-lm:opt2-nosolution}
We will prove it by contradiction. Suppose that $h^*$ is a solution to the optimization problem $\mcP_2$ specified in \eqref{eq:OPT2}. Using the fact that $[J_{\phi_{h^*}}^{-\top}(z)]$ is full-rank for all $z\in\R^n$, and the score difference vector $[s^{\rho_i}(z) - \tilde s^{\rho_i}(z)]$ is not identically zero, \eqref{eq:Dh-entry-z} and Proposition~\ref{prop:continuity-argument} imply that $\bD_{\rm t}(h^*)$ does not have any zero columns. Subsequently, $\norm{\bD_{\rm t}(h^*)}_0 \geq n$, and since $\bD_{\rm t}(h^*)$ is diagonal, we have $\mathds{1}\{\bD_{\rm t}(h^*)\}=\bI_{n \times n}$. We use $J^* \triangleq J_{\phi_{h^*}}^{-\top}$ as shorthand. If $\rho_i = \tilde \rho_i$ for some $i\in[n]$, using $\bD_{\rm t}(h^*)=\bI_{n \times n}$ and Lemma~\ref{lm:parent_change_comprehensive}(ii), for $j\neq i$, we have
\begin{align}\label{eq:no-solution-1}
    0 &= \big[\bD_{\rm t}(h^*)\big]_{\rho_j,\rho_i} \\
    &= \E\bigg[\Big|\big[J^*(Z)]_{\rho_j} \cdot \big[s^{\rho_i}(Z)-\tilde s^{\rho_i}(Z)\big]\Big|\bigg] \\
    &=  \E\bigg[\Big|\big[J^*(Z)]_{\rho_j,i} \cdot \big[s^{\rho_i}(Z) - \tilde s^{\rho_i}(Z)\big]_i \Big|\bigg] \ . 
\end{align}
Recall that, by interventional discrepancy, $[s^{\rho_i}(z) - \tilde s^{\rho_i}(z)]_i \neq 0$ except for a null set. Then, \eqref{eq:no-solution-1} implies that we have $[J^*(z)]_{\rho_j,i} = 0$ except for a null set. Since $J^*$ is continuous, this implies that  $[J^*(z)]_{\rho_j,i} = 0$ for all $z\in \R^n$. Furthermore, since $J^*(z)$ is invertible for all $z$, none of its columns can be a zero vector. Hence, for all $z\in\R^n$,  $[J^*(z)]_{\rho_i,i} = 0$. To summarize, if $\rho_i=\tilde \rho_i$, we have
\begin{align}\label{eq:matching-node-column}
    \forall z\in \R^n \;\; \big[J^*(z)\big]_{j,i} \neq 0 \quad \iff \quad j = \rho_i   \ .
\end{align}
Now, consider the set of \emph{mismatched nodes}
\begin{align}
    \mcA \triangleq \{i \in [n] : \rho_i \neq \tilde \rho_i\} \ .
\end{align}
Let $a \in \mcA$ be a non-descendant of all the other nodes in $\mcA$. There exist nodes $b, c \in \mcA$, not necessarily distinct, such that 
\begin{equation}\label{eq:rho-abc}
    \rho_a = \tilde \rho_b \ , \quad \mbox{and} \quad \rho_c = \tilde \rho_a \ .
\end{equation} 
In four steps, we will show that $\bD(h^*)_{\rho_a,\rho_c}\neq 0$ and $\bD(h^*)_{\rho_c,\rho_a}\neq 0$, which violates the constraint $\mathds{1}\{\bD(h)\} \odot \mathds{1}\{\bD^{\top}(h)\} = \bI_{n \times n}$ and will conclude the proof by contradiction. Before giving the steps, we provide the following argument which we repeatedly use in the rest of the proof. For any continuous function $f:\R^n \to \R$, we have
\begin{align}
    \E\Big[\big|f(Z)\big|\Big] \neq 0 \;\; &\iff \;\; \E\bigg[\Big|f(Z) \cdot \big[s(Z) - s^{\rho_a}(Z)\big]_a \Big|\bigg] \neq 0  \ , \label{eq:continuous-argument-2} \\
    \mbox{and} \quad \E\Big[\big|f(Z)\big|\Big] \neq 0  \;\; &\iff \;\; \E\bigg[\Big|f(Z) \cdot \big[s(Z) - \tilde s^{\rho_c}(Z)\big]_a \Big|\bigg] \neq 0  \ . \label{eq:continuous-argument-3}
\end{align}
First, suppose that $\E\big[|f(Z)|\big] \neq 0$. Then, there exists an open set $\Psi \subseteq \R^n$ for which $f(z)\neq 0$ for all $z\in \Psi$. Due to interventional discrepancy between the pdfs $p_a(z_a)$ and $q_a(z_a)$, there exists an open set within $\Psi$ for which $[s^{\rho_a}(Z) - s(Z)]_a \neq 0$. This implies that
\begin{equation}
    \E\bigg[\Big|f(Z) \cdot \big[s(Z) - s^{\rho_a}(Z)\big]_a \Big|\bigg] \neq 0 \ .
\end{equation}
For the other direction, suppose that $\E\big[\big|f(Z) \cdot [s^{\rho_a}(Z) - s(Z)]_a\big|\big] \neq 0$, which implies that there exists an open set $\Psi$ for which both $f(z)$ and $[s^{\rho_a}(z) - s(z)]_a$ are non-zero. Then, $\E\big[|f(Z)|\big] \neq 0$, and we have \eqref{eq:continuous-argument-2}. Similarly, due to $\rho_c = \tilde \rho_a$ and interventional discrepancy between $p_a$ and $\tilde q_a$, we obtain \eqref{eq:continuous-argument-3}. 

\paragraph{Step 1: Show that $\E\Big[\big|[J^*(Z)]_{\rho_a,a}\big|\Big] \neq 0$.} First, using \eqref{eq:Dh-entry-z} and Lemma~\ref{lm:parent_change_comprehensive}(i), we have
\begin{align}
    \big[\bD(h^*)\big]_{\rho_a,\rho_a} &= \E\bigg[\Big|\big[J^*(Z)\big]_{\rho_a} \cdot \big[s(Z)-s^{\rho_a}(Z)\big]\Big|\bigg] \\ 
    &= \E\bigg[\Big|\sum_{j \in \overline{\Pa}(a)} \big[J^*(Z)\big]_{\rho_a,j} \cdot \big[s(Z)-s^{\rho_a}(Z)\big]_j\Big|\bigg] \ . \label{eq:no-solution-step1}
\end{align}
Note that $\overline{\Pa}(a) \cap \mcA = \{a\}$ since $a$ is non-descendant of the other nodes in $\mcA$. Consider $j\in\Pa(a)$, which implies that $j \notin \mcA$ and $\rho_j=\tilde \rho_j$. By \eqref{eq:matching-node-column}, we have $[J^*(Z)]_{\rho_a,j} = 0$. Then, \eqref{eq:no-solution-step1} becomes
\begin{equation}
    \big[\bD(h^*)\big]_{\rho_a,\rho_a} = \E\bigg[\Big|\big[J^*(Z)\big]_{\rho_a,a} \cdot \big[s(Z)-s^{\rho_a}(Z)\big]_a\Big|\bigg] \neq 0 \ ,
\end{equation} 
since diagonal entries of $\bD(h^*)$ are non-zero due to the last constraint in \eqref{eq:OPT2}. Then, \eqref{eq:continuous-argument-2} implies that $\E\big[\big|[J^*(Z)]_{\rho_a,a}\big|\big] \neq 0$.

\paragraph{Step 2: Show that $\big[\tilde \bD(h^*)\big]_{\rho_a,\rho_c}\neq 0$.} Next, we use $\rho_c=\tilde \rho_a$ and Lemma~\ref{lm:parent_change_comprehensive}(i) to obtain
\begin{align}
    \big[\tilde \bD(h^*)\big]_{\rho_a,\rho_c} &=  \E\bigg[\Big|\big[J^*(Z)\big]_{\rho_a} \cdot \big[s(Z)-\tilde s^{\rho_c}(Z)\big]\Big|\bigg]\\
    &= \E\bigg[\Big|\sum_{j \in \overline{\Pa}(a)} \big[J^*(Z)\big]_{\rho_a,j} \cdot \big[s(Z) - \tilde s^{\rho_c}(Z)\big]_j \Big|\bigg] \\
    &= \E\bigg[\Big|\big[J^*(Z)\big]_{\rho_a,a} \cdot \big[s(Z)-\tilde s^{\rho_c}(Z)\big]_a \Big|\bigg] \ .
\end{align}
Using \eqref{eq:continuous-argument-3} and Step~1 result, we have $\big[\tilde \bD(h^*)\big]_{\rho_a,\rho_c}\neq 0$.

\paragraph{Step 3: Show that $\E\Big[\big|[J^*(Z)]_{\rho_c,a}\big|\Big] \neq 0$.} Using \eqref{eq:Dh-entry-z} and Lemma~\ref{lm:parent_change_comprehensive}(i), we have
\begin{align}
    \big[\tilde\bD(h^*)\big]_{\rho_c,\rho_c} &= \E\bigg[\Big|\big[J^*(Z)\big]_{\rho_c} \cdot \big[s(Z)-\tilde s^{\rho_c}(Z)\big]\Big|\bigg] \\ 
    &= \E\bigg[\Big|\sum_{j \in \overline{\Pa}(a)} \big[J^*(Z)\big]_{\rho_c,j} \cdot \big[s(Z)-\tilde s^{\rho_c}(Z)\big]_j\Big|\bigg] \\ 
    &= \E\bigg[\Big|\big[J^*(Z)]_{\rho_c,a} \cdot \big[s(Z)-\tilde s^{\rho_c}(Z)\big]_a \Big|\bigg] \ .
\end{align}
Since $\mathds{1}\{\bD(h^*)\}=\mathds{1}\{\tilde \bD(h^*)\}$, the diagonal entry $\big[\tilde \bD(h^*)\big]_{\rho_c,\rho_c}$ is non-zero. Then, using \eqref{eq:continuous-argument-3} we have $\E\big[|[J^*(Z)]_{\rho_c,a}|\big] \neq 0$.

\paragraph{Step 4: Show that $\big[\bD(h^*)\big]_{\rho_c,\rho_a}\neq 0$.} Next, we use $\rho_c=\tilde \rho_a$ and Lemma~\ref{lm:parent_change_comprehensive}(i) to obtain
\begin{align}
    \big[\tilde \bD(h^*)\big]_{\rho_c,\rho_a} &=  \E\bigg[\Big|\big[J^*(Z)\big]_{\rho_c} \cdot \big[s(Z)-s^{\rho_a}(Z)\big]\Big|\bigg]\\
    &= \E\bigg[\Big|\sum_{j \in \overline{\Pa}(a)} \big[J^*(Z)\big]_{\rho_c,j} \cdot \big[s(Z) - s^{\rho_a}(Z)\big]_j \Big|\bigg] \\
    &= \E\bigg[\Big|\big[J^*(Z)\big]_{\rho_c,a} \cdot \big[s(Z)- s^{\rho_a}(Z)\big]_a \Big|\bigg] \ .
\end{align}
Using \eqref{eq:continuous-argument-2} and Step~3 result, we have $[\bD(h^*)]_{\rho_c,\rho_a}\neq 0$.

However, using the constraint $\mathds{1}\{\bD(h^*)\}=\mathds{1}\{\tilde \bD(h^*)\}$, we have $[\bD(h^*)]_{\rho_a,\rho_c}\neq 0$. Then, $[\bD(h^*) \odot \bD^{\top}(h^*)]_{\rho_a,\rho_c} \neq 0$, which violates the last constraint in \eqref{eq:OPT2}. Therefore, if the coupling is incorrect, optimization problem $\mcP_2$ has no feasible solution.

\subsection{Proof of Lemma~\ref{lm:opt2-solution}}\label{proof-lm:opt2-solution}
We consider the true encoder $g^{-1}$ under the permutation $\rho^{-1}$, that is $h = \rho^{-1} \circ g^{-1}$, and show that it is a solution to $\mcP_2$ specified in \eqref{eq:OPT2}. First, note that $\phi_{h} = \rho^{-1} \circ g^{-1}  \circ g = \rho^{-1}$, which is just a permutation. Hence, $J_{\phi_{h}}^{-\top}$ becomes the permutation matrix $\bP_{\rho}^{\top}$. Then, for all $i,m \in[n]$ we have 
\begin{align}
    \big[\bD_{\rm t}(h)\big]_{\rho_i,m} &= \E\bigg[\Big|\big[\bP_{\rho}^{\top}\big]_{\rho_i} \cdot \big[s^{m}(Z)-\tilde s^{m}(Z)\big]\Big|\bigg] = \E\bigg[\Big|\big[s^{m}(Z) - \tilde s^{m}(Z)\big]_i\Big|\bigg] \ .
\end{align}
Then, by Lemma~\ref{lm:parent_change_comprehensive}(ii), we have $[\bD_{\rm t}(h)]_{\rho_i,m} \neq 0$ if and only if $i=I^m$, which means $m=\rho_i$ and $\bD_{\rm t}(h)$ is a diagonal matrix. Hence, $h$ satisfies the first constraint.
Next, consider $\bD(h)$. For all $i,j\in[n]$, we have
\begin{align}
    \big[\bD_{\rm t}(h)\big]_{\rho_i,\rho_j} &= \E\bigg[\Big|\big[\bP_{\rho}^{\top}\big]_{\rho_i} \cdot \big[s(Z)-s^{\rho_j}(Z)\big]\Big|\bigg] = \E\bigg[\Big|\big[s^{\rho_j}(Z) - \tilde s^{\rho_j}(Z)\big]_i\Big|\bigg] \ .
\end{align}
By Lemma~\ref{lm:parent_change_comprehensive}(i), we have $[\bD(h)]_{\rho_i,\rho_j} \neq 0$ if and only if $i=\overline{\Pa}(j)$. Since $\rho=\tilde \rho$, similarly, we have $[\tilde D(h_{\rho})]_{\rho_i,\rho_j} \neq 0$ if and only if $i \in \overline{\Pa}(j)$. Therefore, we have $\mathds{1}\{\bD(h)\} = \mathds{1}\{\tilde \bD(h)\}$, $\bD(h)$ has full diagonal and it does not have non-zero values in symmetric entries. Hence, $h$ satisfies the second and third constraints. Therefore, $h$ is a solution to $\mcP_2$ since it satisfies all constraints and the diagonal matrix $\bD_{\rm t}(h)$ has $\norm{\bD_{\rm t}(h)}_0=n$, which is the lower bound established. 

\subsection{Proof of Theorem~\ref{th:two-hard-uncoupled}}\label{appendix-proof-two-hard-uncoupled}

Recall that $\tilde\mcI = \{\tilde I^1,\dots,\tilde I^n\}$ is the permutation of intervened nodes in $\tilde \mcE$, so coupling $\pi$ considered in \eqref{eq:OPT2} is just equal to $\tilde \mcI$. Similarly to the definition of $\rho$ for $\mcI$ in the proof of Theorem~\ref{th:two-hard-coupled}, let $\tilde\rho$ be the permutation that maps $\{1,\dots,n\}$ to $\tilde\mcI$, i.e., $I^{\tilde \rho_i}=i$ for all $i\in[n]$. Then, $\bP_{\tilde \rho}$ denotes the permutation matrix for the intervention order of the environments $\{\tilde \mcE^1,\dots,\tilde \mcE^n\}$, i.e., 
\begin{align}\label{eq:intervention-order-tilde}
    \big[\bP_{\tilde \rho}\big]_{i,j} = \begin{cases}
        1  \ , &j = \tilde \rho_i \ , \\
        0 \ , & \text{else} \ .
    \end{cases}
\end{align}
Lemma~\ref{lm:opt2-no-solution} shows that if the coupling is incorrect, i.e., $\pi \neq \mcI$ or equivalently $\rho \neq \tilde \rho$, the optimization problem in \eqref{eq:OPT2} does not have a feasible solution. Next, Lemma~\ref{lm:opt2-solution} shows that if the coupling is correct, i.e., $\rho=\tilde \rho$, there exists a solution to $\mcP_2$. Lemmas~\ref{lm:opt2-no-solution} and \ref{lm:opt2-solution} collectively prove identifiability as follows. We can search over the permutations of $[n]$ until $\mcP_2$ admits a solution $h^*$. By Lemma~\ref{lm:opt2-no-solution}, the existence of this solution means that coupling is correct. Note that when the coupling is correct, the constraint set of $\mcP_1$ is a subset of the constraints in $\mcP_2$. Furthermore, the minimum value of $\norm{\bD_{\rm t}(h)}_0$ is lower bounded by $n$ (Lemma~\ref{lm:min-score-variations}), which is achieved by the solution $h^*$ (Lemma~\ref{lm:opt2-solution}). Hence, $h^*$ is also a solution to $\mcP_1$, and by Theorem~\ref{th:two-hard-coupled}, it satisfies perfect recovery of the latent DAG and the latent variables.

\subsection{Proof of Proposition~\ref{prop:permutation-diagonal}}\label{proof:permutation-diagonal}
Denote the set of all permutations of $[n]$ by $\mcS_n$. From Leibniz formula for matrix determinants, for a matrix $\bA \in \R^{n \times n}$ we have
\begin{align}
    \det(\bA) &= \sum_{\pi \in \mcS_n} \sgn(\pi) \cdot \prod_{i = 1}^{n} \bA_{i, \pi_i}
\end{align}
where $\sgn(\pi)$ for a permutation $\pi$ of $[n]$ is $+1$ and $-1$ for even and odd permutations, respectively. $\bA$ is invertible if and only if $\det(\bA) \neq 0$, which implies that
\begin{align}
    \exists \pi \in \mcS_n \; : \;\; \sgn(\pi) \cdot \prod_{i = 1}^{n} \bA_{i, \pi_i} \neq 0 \ .
\end{align}
By the definition of $\bP_{\pi}$, we have $[\bP_{\pi} \cdot \bA]_{i, i} = \bA_{i, \pi_{i}}$. Then,
\begin{align}
    \exists \pi \in \mcS_n \; : \;\; \sgn(\pi) \cdot \prod_{i = 1}^{n} [\bP_{\pi} \cdot \bA]_{i, i} \neq 0 \ ,
\end{align}
which holds if and only if $[\bP_{\pi} \cdot \bA]_{i, i} \neq 0$ for all $i \in [n]$.

\section{Details of Simulations}\label{appendix:simulations}

We perform experiments for the coupled environments setting and using a regularized, $\ell_1$-relaxed version of the optimization problem \eqref{eq:OPT1}. Specifically, in Step~2 of GSCALE-I, we solve the following optimization problem:
\begin{equation}\label{eq:OPT3}
    \min_{h \in \mcH}  \|\bD_{\rm t}(h)\|_{1,1} + \lambda_1 \E \| h^{-1}(h(X)) - X \|_2^2 + \lambda_2 \E \| h(X) \|_2^2 \ .
\end{equation}
In this section, we describe data generation, computation of the ground truth score differences for $X$, justification of the optimization problem in \eqref{eq:OPT3} and other implementation details.

\paragraph{Data generation details.}
To generate $\mcG$ we use the Erd{\H o}s-R{\' e}nyi model with $n\in\{5,8\}$ nodes and density $0.5$. For the observational causal mechanisms, we adopt an additive noise model with
\begin{align}
Z_i =\sqrt{Z_{\Pa(i)}^{\top} \cdot \bA_{i} \cdot Z_{\Pa(i)}} + N_{i}\ , \label{eq:quadratic-model--general-appendix}
\end{align}
where $\{\bA_{i}:i\in[n]\}$ are positive-definite matrices, and the noise terms are zero-mean Gaussian variables with variances $\sigma_{i}^2$ sampled randomly from ${\rm Unif}([0.5,1.5])$. We construct the positive-definite matrix $\bA_{i}$ by generating a matrix $\bB_{i} \in \R^{|\Pa(i)| \times |\Pa(i)|}$ by sampling its entries from ${\rm Unif}([0, 1])$ and setting $\bA_{i} = \bB_{i}^{\top} \bB_{i}$.

For two hard interventions on node $i$, $Z_i$ is set to $N_{q,i} \sim \mcN(0,\sigma_{q,i}^2)$ and $N_{\tilde q,i} \sim \mcN(0,\sigma_{\tilde q,i}^2)$. We set $\sigma_{q,i}^2 = \sigma_{i}^2+1$ and $\sigma_{\tilde q,i}^2 = \sigma_{i}^2+2$. We consider target dimension values $d\in\{5,25,40\}$ for $n=5$ and $d\in\{8,25,40\}$ for $n=8$. For each $(n,d)$ pair, we generate 100 latent graphs, and $n_{\rm s}$ samples per environment per graph, where we set $n_{\rm s}=100$ for $n=5$ and $n_{\rm s}=300$ for $n=8$. As the transformation, we consider a generalized linear model,
\begin{align}\label{eq:transf_exp-appendix}
    X &= g(Z) = \tanh (\bG \cdot Z) \ ,
\end{align}
in which $\tanh$ is applied element-wise, and the parameter $\bG \in \R^{d \times n}$ is a randomly sampled full-rank matrix.

\paragraph{Score function of the quadratic model.}
Following \eqref{eq:pz_m_factorized_hard}, score functions $s^m$ and $\tilde s^m$ are decomposed as follows.
\begin{align}
    s^{m}(z) &= \nabla_z \log q_{\ell}(z_{\ell}) +  \sum_{i \neq \ell} \nabla_z \log p_i(z_i \med z_{\Pa(i)}) \ , \label{eq:gscalei_sz_m_decompose_hard} \\
    \mbox{and} \quad \tilde s^{m}(z) &= \nabla_z \log \tilde q_{\ell}(z_{\ell}) +  \sum_{i \neq \ell} \nabla_z \log p_i(z_i \med z_{\Pa(i)})\ . \label{eq:gscalei_sz_m_decompose_hard_tilde}
\end{align}
For additive noise models, the terms in \eqref{eq:gscalei_sz_m_decompose_hard} and \eqref{eq:gscalei_sz_m_decompose_hard_tilde} have closed-form expressions. Specifically, using \eqref{eq:additive-component-score} and denoting the score functions of the noise terms $\{N_i : i \in [n]\}$ by $\{r_i : i \in [n]\}$, we have
\begin{align}
    [s(z)]_i &= r_i(n_i)-\sum_{j \in \Ch(i)} \frac{\partial f_i(z_{\Pa(j)})}{\partial z_i} \cdot r_j(n_j) \ .
\end{align}
In the quadratic latent model we consider in \eqref{eq:quadratic-model--general-experiments}, we have
\begin{align}
    f_i(z_{\Pa(i)}) = \sqrt{z_{\Pa(i)}^{\top} \cdot \bA_i \cdot z_{\Pa(i)}} \ ,
    \quad \mbox{and} \quad
    N_i &\sim \mcN(0, \sigma_i^{2}) \ ,
\end{align}
which implies 
\begin{align}
  \frac{\partial f_j(z_{\Pa(j)})}{\partial z_i} = \frac{[\bA_j]_i \cdot z_{\Pa(j)}}{\sqrt{z_{\Pa(j)}^{\top} \cdot \bA_j \cdot z_{\Pa(j)}}} \ , \quad \mbox{and} \quad r_i(n_i) = -\frac{n_i}{\sigma_i^2} \ , \quad \forall i \in [n] \ .
\end{align}
Components of the score functions $s^m$ and $\tilde s^m$ can be computed similarly. Subsequently, using Corollary~\ref{corollary:sx-from-sz} of Lemma~\ref{lm:score-difference-transform-general}, we can compute the score differences of observed variables as follows.
\begin{align}
    s_X(x) - s_X^m(x) &= \Big[\big[J_g(z)\big]^{\dag}\Big]^{\top} \cdot \big[s(z) - s^m(z)\big] \ , \\
    s_X(x) - \tilde s_X^m(x) &= \Big[\big[J_g(z)\big]^{\dag}\Big]^{\top} \cdot \big[s(z) - \tilde s^m(z)\big] \ , \\   
    s_X^m(x) - \tilde s_X^m(x) &= \Big[\big[J_g(z)\big]^{\dag}\Big]^{\top} \cdot \big[s^m(z) - \tilde s^m(z)\big] \ .   
\end{align}

\paragraph{Implementation and evaluation steps.}

Leveraging \eqref{eq:transf_exp}, we parameterize valid encoders $h$ with parameter $\bH \in \R^{n \times d}$.
\begin{align}
    \hat Z(X;h) &= h(X) = \bH \cdot \arctanh (X) \ , \\
    \hat X &= h^{-1}(\hat Z(X;h)) = \tanh (\bH^{\dag} \cdot \hat Z(X;h)) \ ,
\end{align}
Note that given this parameterization, the function $\phi_{h}(z) = (h \circ g)(z)$ is given by
\begin{equation}
    \hat Z = \phi_{h}(Z) = \bH \cdot \bG \cdot Z \ .
\end{equation}
Subsequently, the only element-wise diffeomorphism between $Z$ and $\hat Z$ is an element-wise scaling. Hence, we can evaluate the estimated latent variables against the scaling consistency metric. To this end, we use normalized $\ell_2$ loss specified in Section~\ref{sec:simulations}.

We use $n_{\rm s}$ samples from the observational environment to compute empirical expectations. Since encoder $h$ is parameterized by $\bH$, we use gradient descent to learn this matrix. To do so, we relax $\ell_0$ norm in \eqref{eq:OPT1} and instead minimize element-wise $\ell_{1,1}$ norm $\|\bD_{\rm t}(h)\|_{1,1}$. Note that, scaling up $\hat Z(h)$ by a constant factor scales down the score differences by the same factor. Hence, to prevent the vanishing of the score difference loss trivially, we add the following regularization term to the optimization objective.
\begin{equation}
    \E\Big[\big\| \hat Z(h) \|_2^2\Big]  = \E\Big[\big\|h(X)\big\|_2^2\Big] \ .
\end{equation}
We also add the following reconstruction loss to ensure that $h$ is an invertible transform.
\begin{equation}
    \E\Big[\big\|h^{-1}(h(X)) - X \big\|_2^2\Big] \ .
\end{equation}
In the end, we minimize the objective function
\begin{equation}\label{eq:l11-loss-w-reg}
    \|\bD_{\rm t}(h)\|_{1,1} + \lambda_1 \E\Big[\big\|h^{-1}(h(X)) - X \big\|_2^2\Big] + \lambda_2 \E\Big[\big\|h(X)\big\|_2^2\Big] \ .
\end{equation}
We denote the final parameter estimate by $\bH^*$ and the encoder parameterized by $\bH^*$ by $h^*$. Note that we do not enforce the diagonality constraint upon $\bD_{\rm t}(h)$. Since we learn the latent variables up to permutation, we change this constraint to a post-processing step. Specifically, we permute the rows of $\bH^*$ to make $\bD_{\rm t}(h^*)$ as close to as diagonal, i.e., $\|\diag ( \bD_{\rm t}(h^*) \odot \bI_n)\|_1$ is maximized.

In the simulation results reported in Section~\ref{sec:simulations}, we set $\lambda_1 = 10^{-4}$ and $\lambda_2 = 1$, and minimize \eqref{eq:l11-loss-w-reg} using RMSprop optimizer with learning rate $10^{-3}$ for $3 \times 10^{4}$ steps for $n=5$ and $4 \times 10^{4}$ steps for $n=8$. Recall that the latent graph estimate $\hat \mcG$ is constructed using $\mathds{1}\{\bD(h^*)\}$. We use a threshold $\lambda_{\mcG}$ to obtain the graph from the upper triangular part of $\bD(h^*)$ as follows.
\begin{equation}
    \hat \Pa(i) = \{j : j < i \quad \mbox{and} \quad [\bD(h^*)]_{j,i} \geq \lambda_{\mcG} \} \ , \qquad \forall i \in [n] \ .
\end{equation}
We set $\lambda_{\mcG}=0.1$ for $n=5$ and $\lambda_{\mcG}=0.2$ for $n=8$.

\end{document}